\documentclass{article}

\usepackage[english]{babel}
\usepackage[letterpaper,top=2cm,bottom=2cm,left=3cm,right=3cm,marginparwidth=1.75cm]{geometry}
\usepackage[table]{xcolor}
\usepackage{amsmath,amssymb}
\usepackage{booktabs}
\usepackage{graphicx}
\usepackage{multirow}
\usepackage{makecell}
\usepackage[normalem]{ulem}
\usepackage{pifont}
\usepackage{tabularx}
\usepackage{wrapfig}
\usepackage{float}
\usepackage[colorlinks=true,allcolors=blue]{hyperref}

\newcommand{\cmark}{\ding{51}}
\newcommand{\xmark}{\ding{55}}
\newcommand{\best}[1]{\textbf{#1}}
\newcommand{\second}[1]{\uline{#1}}

\definecolor{GroupBlue}{RGB}{235,241,249}
\definecolor{GroupGreen}{RGB}{234,246,242}
\definecolor{GroupSand}{RGB}{249,243,235}
\definecolor{GroupSlate}{RGB}{229,239,247}
\definecolor{OursFill}{RGB}{239,247,250}
\definecolor{ProbeGreen}{HTML}{00A67C}
\definecolor{ProbeOrange}{HTML}{D86600}

\title{Enfold: Folding World Model Imagination into Predictive Representations for Ultra-Efficient Embodied Control}
\author{
Weili Zeng\textsuperscript{1}\thanks{Equal contribution.} \quad
Yitong Xing\textsuperscript{1}\footnotemark[1] \quad
Fulong Liu\textsuperscript{1} \quad
Chengqun Yang\textsuperscript{1} \\[0.25em]
Antao Xiang\textsuperscript{1} \quad
Feng Tian\textsuperscript{1} \quad
Jingnan Gao\textsuperscript{1} \quad
Jisong Cai\textsuperscript{1} \\[0.25em]
Xin Wang\textsuperscript{2} \quad
Xiaomin Wu\textsuperscript{2} \quad
Yao Mu\textsuperscript{1} \quad
Xiaokang Yang\textsuperscript{1} \quad
Yichao Yan\textsuperscript{1}\thanks{Corresponding author: \href{mailto:yanyichao@sjtu.edu.cn}{yanyichao@sjtu.edu.cn}.} \\[0.5em]
\textsuperscript{1}Shanghai Jiao Tong University \\
\textsuperscript{2}Qilu University of Technology (Shandong Academy of Sciences) \\[0.35em]
Project page: \url{https://zwl666666.github.io/enfold/}
}
\date{}

\begin{document}
\maketitle

\begin{abstract}
World generative models are typically used through what they produce: a
rendered future, a video-conditioned action, or latent context computed by a
costly generative branch. We argue that their more reusable asset is the
computation that constructs a future. As a generator transforms a corrupted
future into a coherent trajectory, its intermediate states organize appearance,
spatial layout, and interaction across levels of abstraction. Can this
future-generative computation be internalized in a representation inferred
from the present alone?
We present Enfold, which transfers this computation into a representation
predicted from the current visual context and language instruction. During
training, multi-level states exposed as the generator processes the observed
future supervise a current-only encoder. The learned representation is fed back
to condition future generation and is read by task heads without allowing task
gradients to reshape the encoder. At deployment, action prediction no longer
executes the generator. This asymmetry is essential: predicting stochastic,
future-dependent states without access to the future encourages the encoder to
retain transition structure that can be anticipated from the current state,
rather than reproduce every nuisance variation in its teacher.
Across LIBERO, RoboTwin2.0, and real-robot tasks, Enfold supports strong control
while reducing action latency by $3.7\times$ relative to Fast--WAM,
Enfold-Flash reaches $10.1\times$. Representation analyses show that it
suppresses nuisance variation and preferentially captures changes that emerge
over longer horizons. When the current scene is altered by human intervention,
both the generated continuation and the executed actions adapt, which is
inconsistent with fixed trajectory replay. These results recast a world
generator as a source of predictive control representations: its future need
not be materialized at every step if its internal structure can be enfolded into
the present.
\end{abstract}

\section{Introduction}

Effective control requires more than recognizing the current scene: a robot
must anticipate how the scene will change as an interaction unfolds. World
generative models provide a compelling substrate for this capability. Trained
to synthesize future observations, they capture long-range motion, object
interaction, and multiple plausible continuations in observation
space~\cite{nvidia2025cosmoswfm,nvidia2026cosmos3}. This progress has led to
robotic systems that imagine a visual future before acting or jointly model
videos and actions~\cite{du2024video,lv2025f1,bi2025motus,kim2026cosmos}.

\begin{figure}[t]
    \centering
    \includegraphics[
        width=\textwidth
    ]{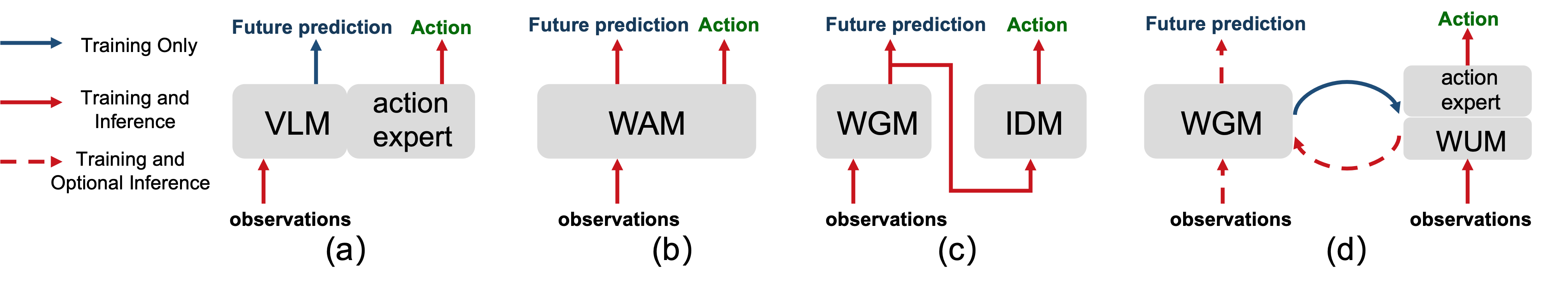}
    \caption{
    \textbf{Different interfaces between future modeling and action prediction.}
    (a) VLA methods augment a vision--language representation
    with future prediction while an action expert produces controls, (b) unified
    WAMs jointly predict future observations and actions, (c) imagine-then-act
    methods generate a visual future with a world generative model (WGM) and
    recover actions through inverse dynamics (IDM), and (d) Enfold instead
    internalizes the WGM's future-generative dynamics in a world understanding
    model (WUM), whose representation conditions both generation and a
    downstream action expert. This representation-level interface avoids
    requiring generated video rollout for action prediction.
    }
    \label{fig:related_work}
\end{figure}

Figure~\ref{fig:related_work} summarizes the dominant output-level interfaces.
Imagine-then-act systems materialize a visual future before recovering an
action, while unified world-action models couple video and action prediction in
one decoding process. Recent work has weakened this dependence on rendered
video: Fast--WAM retains video co-training while skipping explicit future
generation at test time~\cite{yuan2026fastwam}, predictive policies and
asynchronous WAMs consume internal video-model features or reusable latent
context~\cite{hu2025video,cai2026ahawamasynchronoushorizonadaptiveworldactionmodeling},
and ImageWAM reads image-editing caches without decoding the edited
frame~\cite{zhang2026imagewam}. In parallel, VLA--JEPA predicts future
target-encoder features from a current-only pathway~\cite{sun2026vla}. These results show that predictive information need not be carried by a rendered future. They do not, however, reveal whether the computation that constructs such a future can itself be transferred into a representation available from the present alone. We therefore ask: \emph{can the internal computation of a world generator be internalized in a current-only representation that supports both control and future generation?}

The internal states of a video generator are a promising but nontrivial answer.
As a corrupted future is transformed into a coherent trajectory, successive
blocks organize appearance, spatial layout, and interaction structure at
different scales. These states offer a structured alternative to final pixels
as supervision, but they are not ready-made control representations. They
depend on a future unavailable at deployment, vary with the sampled generation
noise, and change substantially across depth and corruption level. A useful
representation must therefore retain the generator's transition-relevant
structure while remaining predictable from the current context.

We call the resulting framework \textbf{Enfold}: it enfolds
future-conditioned generative dynamics into a representation inferred from
the present, from which the model can either unfold a visual future or read out
an action directly. Rather than executing the generator to obtain this
representation at deployment, Enfold trains a separate current-only predictive
encoder. It couples this encoder and a video generator in two directions. In
\emph{generation-to-representation} (G2R), multi-level features exposed as the
generator processes the observed future supervise an encoder that sees only
the current visual context and instruction. In
\emph{representation-to-generation} (R2G), the same representation conditions
future generation. Detached task heads read it without sending task gradients
into the encoder. The generator therefore serves as a training-time supervisor
and an optional future decoder, rather than a mandatory component of the
control loop. At deployment, action prediction executes only the predictive
encoder and action head.

A central design choice is which generator states should supervise the encoder. Our analysis reveals no universally best layer. The most informative block shifts with the corruption level: deeper states increasingly suppress
appearance variation and emphasize global layout, but can also become more
sensitive to generation noise. Enfold therefore predicts a multi-level target
through a timestep-conditioned head. Unlike pixel prediction or distillation
from a frozen visual encoder, this target describes how a generative model
organizes the observed transition across both depth and corruption.

Our results show that this is not merely a cheaper interface. Without embodied pre-training, Enfold obtains $97.8\%$ success on LIBERO and $91.77\%$ on RoboTwin2.0. Enfold-Flash retains $97.5\%$ on LIBERO and reaches $92.02\%$ on
RoboTwin2.0. At $134$\,ms per action chunk, Enfold is $3.7\times$ faster than
Fast--WAM~\cite{yuan2026fastwam}. Enfold-Flash reaches $49$\,ms and is
$10.1\times$ faster.
Generator-state supervision outperforms both future-pixel and action-only
targets, while feeding the learned representation back to the generator
improves future-video prediction. Representation probes further show that it
suppresses nuisance variation present in the generator features and becomes
increasingly predictive of scene regions that change over longer horizons.
Finally, when a human changes the current scene while the goal and preceding
interaction history remain fixed, both the imagined continuation and the
executed actions adapt. This coordinated response is inconsistent with simple
trajectory replay and provides qualitative evidence of counterfactual
consistency under an alternative present.

Our contributions are:
\begin{itemize}
    \item We formulate the multi-level internal computation exposed by a world generator as a predictive supervision space, and
    characterize how its utility and stability vary across depth and corruption.
    \item We develop Enfold, which distills this future-generative computation
    into a current-only representation and makes it a shared interface to future
    generation and stop-gradient task readouts.
    \item We show that this interface supports competitive simulated and
    real-world control without executing the generator in the action path,
    improves future prediction, and substantially reduces action latency.
\end{itemize}

\section{Related Work}

Figure~\ref{fig:related_work} organizes prior work by the interface through
which future information reaches action prediction. The following discussion
reviews three corresponding lines of research: future-aware VLA models,
generative world-action models, and predictive representation learning.

\subsection{Vision--Language--Action Models}

Vision--language--action (VLA) models transfer large-scale vision--language
pre-training to robotic control through scalable adaptation and expressive
action decoding~\cite{brohan2023rt,zitkovich2023rt,kim2025openvla,
black2024pi_0,pi05,openvlaoft,liu2025rdt,zheng2025x,liu2025hybridvla,
chen2025unified,liang2025discrete}. Standard VLA objectives map observations
and instructions directly to actions, leaving environmental dynamics implicit
in action supervision. Recent methods introduce future queries, latent actions,
visual subgoals, or auxiliary prediction to strengthen temporal
reasoning~\cite{zheng2024tracevla,lingbotvla,bu2025univla,hu2025video,
tian2025predictive,yang2025mantis,sun2026vla}. Complementary
efficiency-oriented work predicts feature or cache updates so that an expensive
VLM need not run at every control step~\cite{liu2026latentbridge}.
VLA--JEPA is especially relevant to predictive supervision: a current-only
student predicts target-encoder features computed from future frames, defining
the future in the representation space of a discriminative encoder rather than
through the intermediate computation of a generative model~\cite{sun2026vla}.

\subsection{Generative World-Action Models}
World-action models (WAMs) use predicted scene evolution as supervision or
context for action generation. Imagine-then-act systems first predict a visual
future and recover actions through inverse dynamics or a goal-conditioned
policy~\cite{du2023learning,du2024video,black202susie,hatch2024ghilglue,
bu2024clover,tian2025predictive,hu2025video}. Unified architectures instead
model videos and actions jointly, allowing dense visual dynamics to complement
action labels~\cite{lv2025f1,bi2025motus,kim2026cosmos,li2026lingbotva,
ye2026world,team2026beingh07,zhang2026lingbotva2,nvidia2026cosmos3}. Other
work reduces explicit rollout through auxiliary future objectives, reusable
latent context, visual subgoals, asynchronous prediction, or structured world
representations~\cite{yuan2026fastwam,gao2025adaworld,team2026beingh07,
cai2026ahawamasynchronoushorizonadaptiveworldactionmodeling,
zhang2026imagewam,yu2026maskwam,zhang2026r3dp}.
Among these, VPP conditions a policy on predictive features extracted inside a
video diffusion model~\cite{hu2025video}. Fast--WAM retains video co-training
but removes explicit future generation from its action-time
path~\cite{yuan2026fastwam}. AHA--WAM amortizes a low-frequency video branch through
reusable layer-wise context, whereas ImageWAM supplies its action expert with
the caches produced by image-editing
denoising~\cite{cai2026ahawamasynchronoushorizonadaptiveworldactionmodeling,
zhang2026imagewam}. These methods establish that useful world-model context need
not take the form of decoded video.

\subsection{Predictive Representation Learning}
Predictive representation learning models temporal structure in feature space
rather than reconstructing complete observations. JEPA-style methods predict
future target-encoder features, while latent-action methods infer transition
variables through inverse and forward dynamics~\cite{assran2023ijepa,
bardes2023vjepa,assran2025vjepa2selfsupervisedvideo,schmidt2024lapo,
bruce2024genie,ye2024lapa}. Pre-trained visual features have also been used as
latent dynamics spaces for planning~\cite{zhou2024dinowm}. These
representations support VLA pre-training, transfer, and alignment with
executable robot actions~\cite{chen2024igor,bu2025univla,
chen2025motolatentmotiontoken,chen2025villaxenhancinglatentaction}. Most such
targets are produced by an encoder or an inferred transition variable and
therefore describe future observations without exposing the intermediate
temporal computation of a generative model.

\paragraph{Positioning of this work.}
Enfold differs from prior work in where predictive information is obtained and how it enters control. VLA--JEPA predicts features defined by a future encoder, whereas VPP, AHA--WAM, and ImageWAM obtain action-time context from an executed generative path or its caches. Fast--WAM avoids video rollout at inference but learns video and action prediction within a coupled architecture. Enfold instead uses multi-level states from a teacher-forced generator as training targets for a current-only encoder. The resulting representation is not shaped by task gradients, can condition both action prediction and future generation, and requires no generator execution in the control path. Enfold therefore transfers future-conditioned generative computation into the present representation, rather than retaining the generator as part of action-time computation.

\section{Motivation: Why Generator States?}
\label{sec:predictive_targets}
Transferring generative computation into a current-only representation requires deciding what the encoder should predict. Action supervision provides executable targets, but only indirect evidence of how the scene will evolve. Future pixels provide dense temporal supervision, yet require the representation to preserve every visual detail, including appearance variation that may be irrelevant to control. Features from a frozen future encoder are more compact, but describe the future as an encoded endpoint rather than exposing the computation through which a generative model constructs it.

Generator states offer a different supervision space. During a teacher-forced pass, the generator processes an observed future across a sequence of corruption levels and network depths. Its intermediate states expose how appearance, spatial configuration, and interaction structure are progressively organized into a coherent trajectory. They are therefore grounded in the realized transition, but need not assign equal importance to every output pixel. If this computation can be predicted from the current context, it provides a route to acquiring future-oriented structure without executing the generator at deployment.

These states are not ready-made targets. They depend on the realized future and
sampled noise, and their content changes across the generator. A useful target
must therefore be informative, predictable enough from the present, and
non-degenerate. We examine this trade-off before defining Enfold.

Let $c$ denote the observed video context, $e$ the language instruction, $y$
the realized future video chunk, and $y_t$ its corrupted latent at timestep
$t$. For this diagnostic, we write the generator before introducing Enfold's
representation conditioning. A teacher-forced forward pass exposes the
hierarchy
\begin{equation}
\mathcal{H}_{\theta}(y_t,t,c,e)
=
\left\{
h^{G}_{\ell,t}
\right\}_{\ell\in\mathcal{L}}.
\label{eq:generator_hierarchy}
\end{equation}
We evaluate task and scene retrieval, effective rank, sensitivity to sampled
noise, and controlled responses to illumination and spatial layout. Together,
these diagnostics measure utility, stability, selectivity, and feature
diversity. Appendix~\ref{app:generator_probe} provides the complete protocol.

\begin{figure}[htp]
    \centering
    \includegraphics[
        width=0.85\textwidth
    ]{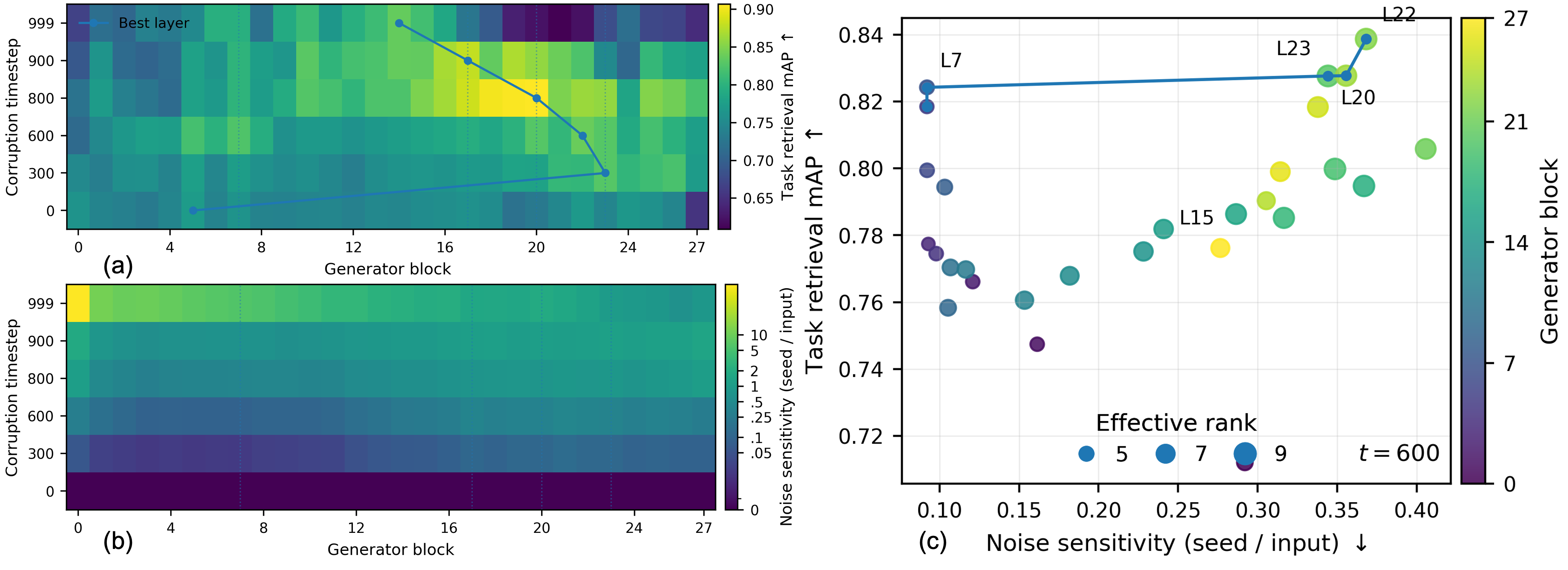}
    \vspace{-3mm}
    \caption{
    \textbf{Predictive utility and stability of generator states.}
    \textbf{(a)} Task/scene retrieval across generator blocks and corruption
    timesteps. The trajectory marks the best-performing block, and dotted lines
    indicate the selected supervision layers.
    \textbf{(b)} Sensitivity to generation noise relative to input variation.
    \textbf{(c)} Utility--stability trade-off at $t=600$; marker size denotes
    effective rank and the curve marks the Pareto frontier.
    }
    \label{fig:generator_target_hierarchy}
\end{figure}

Figure~\ref{fig:generator_target_hierarchy} shows that no block dominates
throughout the corruption trajectory. The most discriminative block changes
with $t$, and depths occupy different points on the utility--stability
frontier. Some late states remain semantically informative while becoming
sensitive to sampled noise, making semantic utility alone insufficient for
selecting a supervision layer.

\begin{figure}[htp]
    \centering
    \includegraphics[
        width=0.85\textwidth
    ]{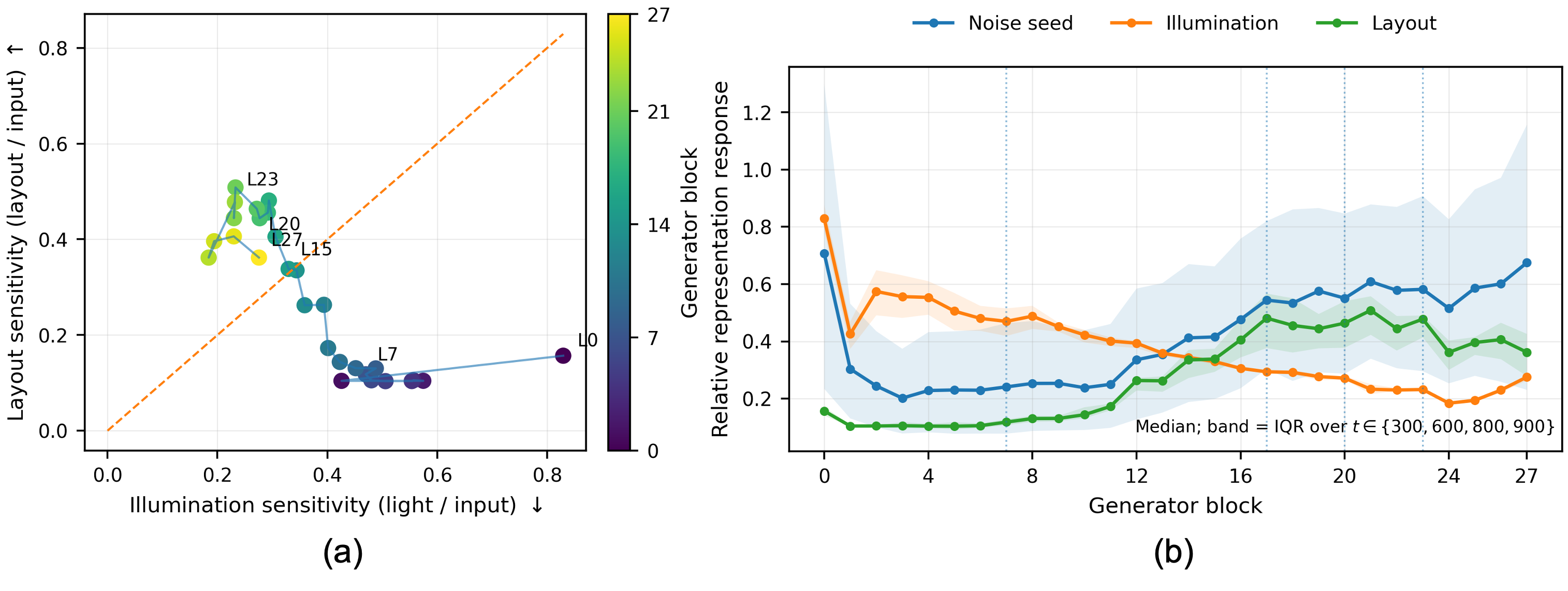}
    \vspace{-6mm}
    \caption{
    \textbf{Depth-dependent selective invariance of generator states.}
    \textbf{(a)} Illumination and layout sensitivity across blocks, summarized
    over $t\in\{300,600,800,900\}$.
    \textbf{(b)} Sensitivity to noise, illumination, and layout across depth.
    Curves and bands denote the median and interquartile range; dotted lines
    mark the selected supervision layers.
    }
    \label{fig:generator_target_selectivity}
\end{figure}

Depth also changes what information is preserved.
Figure~\ref{fig:generator_target_selectivity} shows that deeper states
increasingly suppress illumination changes while remaining responsive to
task-relevant layout, yet can also become more coupled to stochastic
generation. Selectivity and predictability therefore do not improve
monotonically together; different levels expose complementary aspects of the
transition.

These observations motivate a target constructed from several generator
depths $
A\left(
\mathcal{H}_{\theta}(y_t,t,c,e)
\right)$,
where $A(\cdot)$ selects and concatenates the chosen states. Concatenation
preserves complementary levels, while a timestep-conditioned head accounts for
how their structure is expressed at each $t$. The following section
incorporates these choices into the G2R and R2G objectives.

\section{Method}

\begin{figure}[t]
    \centering
    \includegraphics[
        width=0.9\textwidth
    ]{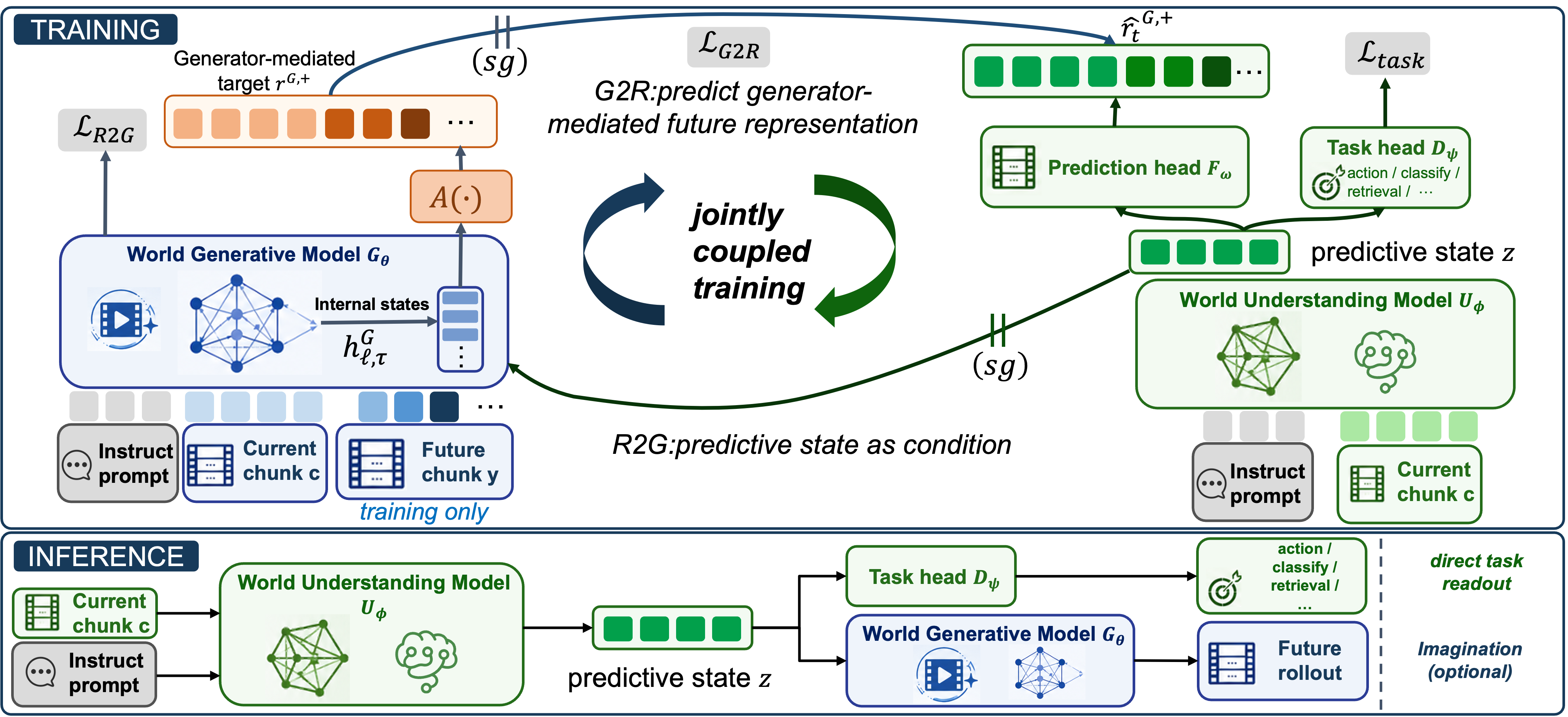}
    \caption{
\textbf{Overview of Enfold.}
The predictive encoder $U_{\phi}$ maps the observed context and instruction to
$z$. In G2R, the generator $G_{\theta}$ processes the corrupted real future.
selected hidden states form a multi-level target predicted from $z$ by the
timestep-conditioned head $F_{\omega}$. In R2G, a detached copy of $z$
conditions future generation. Task heads also read a detached $z$, so task
supervision does not directly shape the encoder. At control time, only $U_{\phi}$ and the action head are executed. $G_{\theta}$ is used only when an
explicit video future is requested.
}
\label{fig:enfold_overview}
\end{figure}

\subsection{Overview}

Enfold contains a predictive encoder $U_{\phi}$, a generative world model
$G_{\theta}$, and lightweight prediction and task heads. For a current video
context $c$ and language instruction $e$, the encoder produces
\begin{equation}
z=U_{\phi}(c,e),
\qquad
\bar z=\operatorname{sg}(z),
\label{eq:problem_setup}
\end{equation}
where $\operatorname{sg}$ denotes stop-gradient. The representation complements
rather than replaces the visual context. G2R determines which future-conditioned
dynamics are enfolded into $z$, R2G tests whether they can be unfolded into a
future. Task heads evaluate the same representation without sending task
gradients into $U_{\phi}$. The two principal paths are coupled in their forward
computation but separated in optimization.

\subsection{Generator-to-Representation Learning}

Let $y_0$ be the latent encoding of the real future chunk and
$\epsilon\sim\mathcal{N}(0,I)$. At timestep $t$, we construct
$y_t=\alpha_t y_0+\sigma_t\epsilon$. While processing $y_t$ under the current
context, instruction, and detached representation, the generator produces
\begin{equation}
\mathcal{H}_{\theta}
\bigl(
y_t,t,c,e,\bar z
\bigr)
=
\left\{
h^{G}_{\ell,t}
\right\}_{\ell\in\mathcal{L}},
\qquad
r^{G}_t
=
\operatorname{LN}
\left[
A\left(
\mathcal{H}_{\theta}
\bigl(
y_t,t,c,e,\bar z
\bigr)
\right)
\right],
\label{eq:teacher_target}
\end{equation}
where $\mathcal{L}$ contains the selected layers and $A(\cdot)$ concatenates
their spatio-temporal tokens along the channel dimension. Layer normalization
places the combined target on a stable scale.

This target is extracted from a teacher-forced forward pass over the real
future, not from a sampled video rollout. It captures how the generator
organizes the observed transition at the sampled corruption level. Because
$z$ itself does not depend on $t$, a timestep-conditioned head maps the shared
representation to the generator's timestep-specific coordinate system:
\begin{equation}
\hat r^{G}_t
=
F_{\omega}\bigl(z,\gamma(t)\bigr),
\qquad
\mathcal{L}_{\mathrm{G2R}}
=
\operatorname{SmoothL1}
\left(
\hat r^{G}_t,
\operatorname{sg}(r^{G}_t)
\right),
\label{eq:g2r}
\end{equation}
where $\gamma(t)$ is the timestep embedding. The same $t$ corrupts the future,
indexes the teacher features, and conditions $F_{\omega}$. Gradients through
the target are stopped. Thus $U_{\phi}$ must predict future-dependent structure
from the present, while $F_{\omega}$ absorbs variation specific to the
generative trajectory.

\subsection{Representation-to-Generation Learning}

We instantiate $G_{\theta}$ as a conditional latent video generator and write
the objective using flow matching~\cite{lipman2023flowmatching}. The generator
predicts the velocity field of the corrupted future conditioned on $c$, $e$,
and $\bar z$:
\begin{equation}
\mathcal{L}_{\mathrm{R2G}}
=
\mathbb{E}_{c,e,y_0,\epsilon,t}
\left[
\left\|
v_{\theta}
\bigl(
y_t,t,c,e,\bar z
\bigr)
-
u_t(y_0,\epsilon)
\right\|_2^2
\right],
\label{eq:r2g}
\end{equation}
where $u_t$ is the target velocity of the interpolation path. Detaching $z$
ensures that this objective teaches the generator to use the representation
without allowing the generation loss to shape the encoder directly. R2G is
therefore also a functional test: if $z$ contains predictive information
beyond the visible context, conditioning on it should improve future modeling.

\subsection{Detached Task Readouts}

Given a task target $q$, a readout $D_{\psi}$ is optimized from the detached
representation:
\begin{equation}
\hat q
=
D_{\psi}\bigl(\operatorname{sg}(z)\bigr),
\qquad
\mathcal{L}_{\mathrm{task}}
=
\ell_{\mathrm{task}}(\hat q,q).
\label{eq:task_readout}
\end{equation}
For embodied tasks, the action head is implemented as a conditional
flow-matching model: it learns an action velocity field conditioned on
$\operatorname{sg}(z)$ and generates action chunks by transporting Gaussian
noise toward the demonstrated action distribution. Detachment makes downstream
performance a readout of the information exposed by generator-mediated
learning, rather than the result of action gradients adapting $U_{\phi}$ into a
policy backbone.

\subsection{Optimization and Inference}

The complete objective is
\begin{equation}
\mathcal{L}
=
\mathcal{L}_{\mathrm{G2R}}
+
\lambda_{\mathrm{R2G}}\mathcal{L}_{\mathrm{R2G}}
+
\lambda_{\mathrm{task}}\mathcal{L}_{\mathrm{task}},
\label{eq:full_objective}
\end{equation}
with the following gradient routes:
\begin{equation}
\begin{aligned}
\nabla_{\phi,\omega}\mathcal{L}
&=
\nabla_{\phi,\omega}\mathcal{L}_{\mathrm{G2R}}, &
\nabla_{\theta}\mathcal{L}
&=
\lambda_{\mathrm{R2G}}
\nabla_{\theta}\mathcal{L}_{\mathrm{R2G}}, &
\nabla_{\psi}\mathcal{L}
&=
\lambda_{\mathrm{task}}
\nabla_{\psi}\mathcal{L}_{\mathrm{task}}.
\end{aligned}
\label{eq:gradient_flow}
\end{equation}

The generator is updated only by future modeling, the predictive encoder and
prediction head only by G2R, and each task head only by its readout loss.

At inference, $U_{\phi}$ is evaluated once to obtain $z$. Future imagination
uses $G_{\theta}$ to sample from $p_{\theta}(y\mid c,e,z)$. Control instead
generates an action chunk directly from $z$ and never executes the video
generator. This representation-level separation provides the algorithmic
latency reduction. Enfold-Flash preserves the same model and inference path and
adds TensorRT operator-level acceleration.

\section{Experiments}
\label{sec:experiments}

The evaluation follows the information flow in
Figure~\ref{fig:enfold_overview} and asks four questions. First, can the
predictive representation support accurate control without executing the video
generator, and what latency does this interface achieve? Second, does it adapt
to an intervened scene, or merely replay a trajectory associated with the
preceding context? Third, do the G2R and R2G paths contribute in their intended
directions? Finally, what information does generator-mediated supervision add
beyond a frozen visual representation? We address these questions in order,
moving from task-level utility and behavior to mechanism and representation.

\subsection{Experimental Setup}
\label{sec:experimental_setup}

We evaluate control on the $40$ language-conditioned tasks drawn from
LIBERO~\cite{liu2023libero}, the $50$ dual-arm tasks of
RoboTwin2.0~\cite{chen2025robotwin2}, and four real-world bimanual tasks.
RoboTwin2.0 is tested in both clean and randomized scenes. Enfold instantiates
the generator with Cosmos-Predict 2.5 2B~\cite{nvidia2025cosmoswfm} and the predictive encoder
with DINOv3 ViT-H+/16~\cite{simeoni2025dinov3}. Simulation control is measured
by task success rate, and real-robot evaluation uses a normalized completion score
over ordered task stages. Future prediction uses PSNR,
SSIM~\cite{wang2004ssim}, and LPIPS~\cite{zhang2018lpips}. Representation
quality is studied with frozen probes and direct feature comparisons. Training,
data processing, action-head configurations, and all evaluation protocols are
detailed in
Appendix~\ref{app:experimental_details}.

\subsection{Efficient Control without Video Rollout}
\label{sec:actionable_representation}

We first test the intended deployment mode: the action head reads $z$ directly,
and the video generator is absent from the control path. The central comparison
is therefore not accuracy alone, but the trade-off between task success and the
cost of exposing future-oriented information to the policy.

\paragraph{LIBERO.}

\begin{table}[t]
\centering
\caption{Comparison with representative vision--language--action and world-action models on the LIBERO benchmark. ``P.T.'' indicates whether embodied pre-training is used. Model size denotes the total number of parameters. All latency values are measured per action prediction on the same NVIDIA A100 40GB GPU. Enfold-Flash uses TensorRT~\cite{nvidia2026tensorrt} for operator-level acceleration. The best value(s) are shown in \best{bold}, and the second-best distinct value(s) are \second{underlined}.}
\label{tab:libero_main_comparison}
\small
\setlength{\tabcolsep}{4.2pt}
\renewcommand{\arraystretch}{1.08}

\begin{tabular}{@{}lcccccccc@{}}
\toprule
\textbf{Method} & \textbf{P.T.} & \makecell[c]{\textbf{Model}\\\textbf{Size}} & \makecell[c]{\textbf{Latency$\downarrow$}\\\textbf{(ms)}} & \textbf{Long} & \textbf{Goal} & \textbf{Object} & \textbf{Spatial} & \textbf{Average} \\
\midrule

\rowcolor{GroupBlue}
\multicolumn{9}{c}{\textsc{Mainstream Vision--Language--Action Models}} \\
OpenVLA--OFT~\cite{openvlaoft} & \cmark & 7B & --- & 94.5 & 97.9 & 98.4 & 97.6 & 97.1 \\
$\pi_{0}$~\cite{black2024pi_0} & \cmark & 3.5B & 184 & 88.4 & 94.4 & 96.8 & 98.0 & 94.4 \\
$\pi_{0.5}$~\cite{pi05} & \cmark & 3.5B & 184 & 92.4 & \second{98.0} & 98.2 & \best{98.8} & 96.9 \\
GR00T--N1.6~\cite{groot_n16} & \cmark & 3.3B & 232 & 94.4 & 97.5 & 98.5 & 97.7 & 97.0 \\

\rowcolor{GroupGreen}
\multicolumn{9}{c}{\textsc{Latent Action-Based Methods}} \\
LAPA~\cite{ye2024lapa} & \cmark & 7B & --- & 55.4 & 58.8 & 74.6 & 73.8 & 65.7 \\
UniVLA~\cite{bu2025univla} & \cmark & 7B & --- & 92.0 & 95.6 & 96.8 & 96.5 & 95.2 \\
Mantis~\cite{yang2025mantis} & \cmark & 5.8B & --- & 94.2 & 94.4 & 99.2 & \best{98.8} & 96.7 \\
VLA--JEPA~\cite{sun2026vla} & \xmark & 3B & --- & 95.8 & 97.2 & 99.6 & 96.2 & 97.2 \\

\rowcolor{GroupSand}
\multicolumn{9}{c}{\textsc{World Action Models}} \\
F1~\cite{lv2025f1} & \xmark & 4B & 352 & 91.3 & 95.4 & 97.8 & 98.2 & 95.7 \\
Motus~\cite{bi2025motus} & \xmark & 8B & 2759 & \second{97.6} & 96.6 & \second{99.8} & 96.8 & 97.7 \\
Cosmos--Policy~\cite{kim2026cosmos} & \xmark & 2.1B & 1133 & \second{97.6} & \best{98.2} & \best{100.0} & 98.1 & \best{98.5} \\
LingBot--VA~\cite{li2026lingbotva} & \xmark & 5.5B & 3812 & \best{98.5} & 97.2 & 99.6 & \second{98.5} & \best{98.5} \\
Fast--WAM~\cite{yuan2026fastwam} & \xmark & 6B & 493 & 95.2 & 97.0 & \best{100.0} & 98.2 & 97.6 \\

% \rowcolor{GroupSlate}
% \multicolumn{9}{c}{\textsc{Representation-Level Interface}} \\
\rowcolor{OursFill}
\textbf{Enfold (Ours)} & \xmark & 3B & \second{134} & 97.4 & 96.8 & \best{100.0} & 97.0 & \second{97.8} \\
\rowcolor{OursFill}
\textbf{Enfold-Flash (Ours)} & \xmark & 3B & \best{49} & 96.6 & 96.6 & 99.8 & 97.0 & 97.5 \\
\bottomrule
\end{tabular}
\end{table}

Table~\ref{tab:libero_main_comparison} reports performance on the four LIBERO
suites. Enfold reaches $97.8\%$ average success without embodied pre-training, including
$100.0\%$ on LIBERO-Object and $97.4\%$ on LIBERO-Long. Relative to Fast--WAM,
it improves the average by $0.2$ points and LIBERO-Long by $2.2$ points, while
using half as many parameters and predicting an action in $134$\,ms---a
$3.7\times$ speedup under the same A100 40GB protocol. The task gains are
modest in an already saturated benchmark. The more consequential result is
that removing generative rollout does not sacrifice overall control quality.

Enfold-Flash further reduces latency to $49$\,ms, or $10.1\times$ faster than
Fast--WAM, with only a $0.3$-point reduction in success. Standard Enfold and
Enfold-Flash separate two sources of efficiency. The former obtains its
$134$\,ms latency from the representation-level interface: video modeling
shapes $z$ during training, but control never executes a future-video rollout.
The latter adds operator-level acceleration without changing that interface.

\paragraph{RoboTwin2.0.}
RoboTwin2.0 tests the same design on dual-arm manipulation and includes a
randomized evaluation protocol.

Enfold obtains $91.60\%$ in the clean setting and $91.94\%$ under
randomization, for a $91.77\%$ average
(Table~\ref{tab:robotwin_results}). Enfold-Flash reaches $91.96\%$ in clean
scenes and $92.08\%$ under randomization, yielding a $92.02\%$ average. This is
$0.19$ points above Fast--WAM ($91.83\%$) and $0.18$ points below LingBot--VA
($92.20\%$), despite using no embodied pre-training. These margins are too
small to support a ranking claim without uncertainty estimates.

\begin{wraptable}[12]{R}{0.52\textwidth}
\centering
\caption{\small Results on RoboTwin2.0. Best results are shown in
\best{bold}, and second-best distinct results are \second{underlined}.}
\label{tab:robotwin_results}
\scriptsize
\setlength{\tabcolsep}{2.4pt}
\renewcommand{\arraystretch}{0.98}
\begin{tabular}{@{}l|c|ccc@{}}
\toprule
\textbf{Method} & \textbf{P.T.} &
\textbf{Clean} & \textbf{Rand.} & \textbf{Avg.} \\
\midrule
$\pi_{0}$~\cite{black2024pi_0}
& \cmark & 65.92 & 58.40 & 62.16 \\
$\pi_{0.5}$~\cite{pi05}
& \cmark & 82.74 & 76.76 & 79.75 \\
ABot--M0~\cite{abot_m0}
& \xmark & 81.20 & 80.40 & 80.80 \\
Motus~\cite{bi2025motus}
& \cmark & 88.66 & 87.02 & 87.80 \\
LingBot--VA~\cite{li2026lingbotva}
& \cmark & \best{92.90} & 91.50 & \best{92.20} \\
Fast--WAM~\cite{yuan2026fastwam}
& \xmark & 91.88 & 91.78 & 91.83 \\
\rowcolor{OursFill}
\textbf{Enfold (Ours)}
& \xmark & 91.60 & \second{91.94} & 91.77 \\
\rowcolor{OursFill}
\textbf{Enfold-Flash (Ours)}
& \xmark & \second{91.96} & \best{92.08} & \second{92.02} \\
\bottomrule
\end{tabular}
\end{wraptable}

The relevant result is instead that TensorRT acceleration preserves control quality:
Enfold-Flash slightly improves the average over Enfold, while both variants
retain performance under the randomized protocol. This shows that the
representation-level interface transfers from single-arm LIBERO to dual-arm
manipulation without requiring video rollout at control time.

\paragraph{Real-world bimanual manipulation.}
We evaluate the same representation-driven policy on four AgileX dual-arm
tasks.
The primary evaluation measures in-distribution task completion. A second split
introduces controlled changes to object placement and scene appearance. This
separates basic task execution from robustness to deviations not present in the
demonstrations.

\begin{table}[H]
\centering
\small
\caption{\textbf{Real-robot normalized completion score (\%).} ID and OOD
denote in- and out-of-distribution evaluation; the stage-based scoring protocol
is detailed in Appendix~\ref{app:experimental_details}.}
\label{tab:real_robot_results}
\setlength{\tabcolsep}{4.0pt}
\renewcommand{\arraystretch}{0.96}
\begin{tabular}{lcccccc}
\toprule
Method & Fold Towel & Organize Desktop & Spoon Powder & Store Plate
& ID Avg. & OOD Avg. \\
\midrule
Motus~\cite{bi2025motus} & 50.0 & 66.7 & 33.3 & 53.3 & 50.8 & 38.0 \\
$\pi_{0.5}$~\cite{pi05} & \best{93.3} & 84.4 & 85.6 & 81.1 & 86.1 & \best{83.3} \\
Fast--WAM~\cite{yuan2026fastwam} & 73.3 & \best{88.9} & 72.2 & 76.7 & 77.8 & 70.0 \\
\rowcolor{OursFill}
Enfold            & 92.2 & \best{88.9} & \best{87.7} & \best{90.0} & \best{89.7} & 76.6 \\
\rowcolor{OursFill}
Enfold-Flash      & 84.4 & 82.2 & 84.4 & 88.9 & 85.0 & 73.3 \\
\bottomrule
\end{tabular}
\end{table}

Enfold reaches an average normalized completion score of $89.7\%$ in
distribution, $11.9$ points above Fast--WAM and $3.6$ points above
$\pi_{0.5}$. The improvement over Fast--WAM is not confined to one task: it is
$18.9$ points on \textit{Fold Towel}, $15.5$ on \textit{Spoon Powder}, and
$13.3$ on \textit{Store Plate}, while the two methods tie on
\textit{Organize Desktop}. This pattern is consistent with the predictive
representation helping most when actions must track deformable geometry or an
evolving object configuration.

Under distribution shift, $\pi_{0.5}$ obtains the strongest OOD aggregate at
$83.3\%$. Enfold ranks second at $76.6\%$, $6.6$ points above Fast--WAM.
Enfold-Flash records normalized scores of $85.0\%$ in distribution and
$73.3\%$ OOD, retaining gains of $7.2$ and $3.3$ points over Fast--WAM,
respectively. Thus, Enfold's advantage over Fast--WAM extends beyond the
demonstrated task distribution, while the $6.7$-point gap to $\pi_{0.5}$
identifies robustness under larger distribution shifts as an important
remaining challenge.

\subsection{Reimagining and Recovery after Intervention}
\label{sec:intervention_recovery}

Aggregate success does not reveal whether Enfold tracks the current scene or
replays a trajectory associated with the preceding context. We intervene on
\textit{Store Plate} and \textit{Fold Towel} at $t_2$, changing the object
configuration while keeping the instruction and prior interaction fixed.
Enfold then re-encodes the scene into $z^{+}$. A replay-based model should retain
the now-invalid continuation, while a context-sensitive model should revise both its
imagined future and its actions.

\begin{figure}[H]
    \centering
    \includegraphics[width=0.98\textwidth]{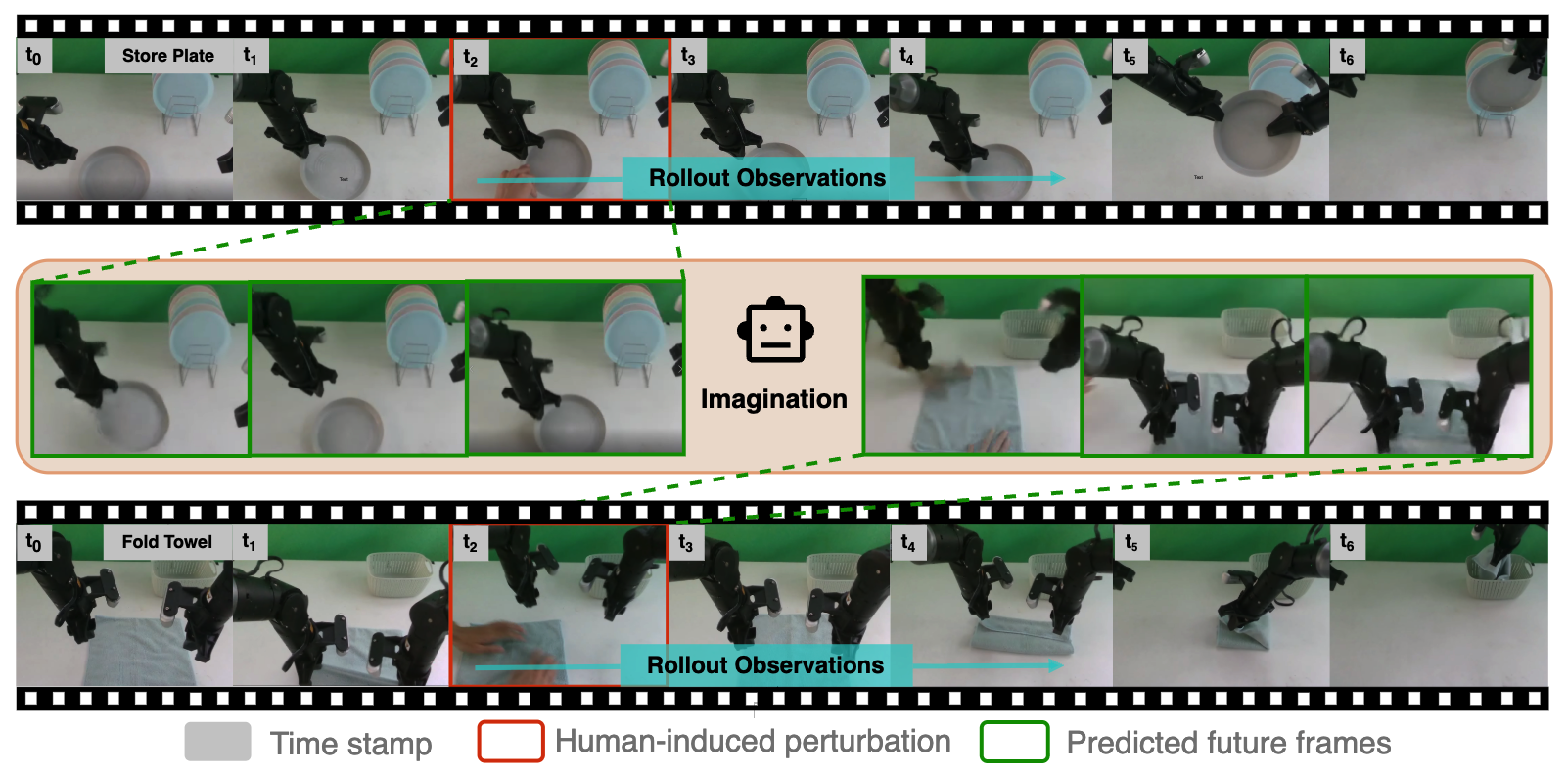}
    \caption{
    \textbf{Reimagining and recovery under human intervention.}
    Red marks the intervention at $t_2$. Green shows future frames decoded from
    the updated representation for \textit{Store Plate} (left) and
    \textit{Fold Towel} (right). The subsequent rollouts adapt to the modified
    scene and complete the original instruction. The generator is used only
    for visualization. Actions are predicted directly from the representation.
    }
    \label{fig:perturbation_recovery}
\end{figure}

In \textit{Store Plate}, the imagined and executed continuations redirect
toward the displaced plate. In \textit{Fold Towel}, they adapt to the changed
cloth geometry and re-establish contact. The important observation is their
coordinated departure from the pre-intervention trajectory. This behavior
argues against fixed trajectory replay and provides qualitative evidence of
counterfactual consistency: Enfold constructs a goal-compatible continuation
from an alternative present. Because the perturbed scene is directly observed,
the result establishes intervention-conditioned replanning rather than full
causal counterfactual inference.

\subsection{Dissecting the Bidirectional Coupling}
\label{sec:why_enfold_works}

The main results establish that $z$ supports control without video rollout, but
do not by themselves show that the generator-mediated objective is responsible.
We therefore test the two directional claims separately. G2R should make
generator states a better supervision target for control than pixels or actions
alone. R2G should make the resulting representation useful beyond the action head,
including to the generator that supplied its supervision.

\begin{wraptable}{r}{0.52\columnwidth}
    \centering
    \vspace{-24pt}
    \caption{\scriptsize Effect of representation supervision on LIBERO (\%).}
    \label{tab:target_ablation}
    \tiny
    \setlength{\tabcolsep}{3pt}
    \renewcommand{\arraystretch}{0.92}
    \resizebox{\linewidth}{!}{
    \begin{tabular}{lccccc}
        \toprule
        Supervision & Spa. & Obj. & Goal & L10 & Avg. \\
        \midrule
        Future pixels
        & \textbf{98.0} & 99.8 & 91.8 & 94.2 & 96.0 \\
        Action only
        & 97.6 & 99.6 & 92.8 & 89.6 & 94.9 \\
        Single-level states
        & 96.8 & 99.8 & 94.0 & 94.6 & 96.3 \\
        Multi-level states
        & 97.0 & \textbf{100.0} & \textbf{96.8}
        & \textbf{97.4} & \textbf{97.8} \\
        \bottomrule
    \end{tabular}
    }
    \vspace{-5pt}
\end{wraptable}

\paragraph{G2R: generator states as supervision.}
We replace the generator-mediated target with either future-pixel prediction or
action-only training, leaving the downstream protocol unchanged.
Table~\ref{tab:target_ablation} separates the value of generator-state
supervision from the value of combining generator depths. A single-level target
reaches $96.3\%$, modestly improving over future-pixel prediction ($96.0\%$)
and action-only training ($94.9\%$). Multi-level supervision raises the average
to $97.8\%$, a further $1.5$-point gain. This gain is concentrated on
LIBERO-Goal and LIBERO-10, where the multi-level target improves over its
single-level counterpart by $2.8$ points on each suite, but is only $0.2$ points
better on LIBERO-Spatial and LIBERO-Object. The pattern supports the use of
complementary generator depths rather than treating one hidden level as a
sufficient summary, particularly when success depends on goal-directed or
multi-stage interaction. Future pixels remain strongest on LIBERO-Spatial by
$1.0$ point, consistent with precise appearance and geometry being sufficient
for more local spatial rearrangements. Because the downstream protocol is
unchanged, these differences isolate the supervision target rather than the
action-head architecture.

\paragraph{R2G: conditioning the generator on the learned representation.}
\label{sec:world_modeling_ability}
We compare Enfold with a video-only Cosmos baseline under a matched training
budget and evaluation protocol. Both models predict the future from the same
visual and language context.

\begin{wraptable}{r}{0.4\linewidth}
    \centering
    \vspace{-8pt}
    \small
    \setlength{\tabcolsep}{4pt}
    \caption{Effect of R2G conditioning on LIBERO future prediction.}
    \label{tab:world_modeling_ability}
    \begin{tabular}{lcc}
        \toprule
        Metric & Video-only & Enfold \\
        \midrule
        PSNR $\uparrow$
        & $25.92$ & $\mathbf{27.27}$ \\
        SSIM $\uparrow$
        & $0.885$ & $\mathbf{0.902}$ \\
        LPIPS $\downarrow$
        & $0.0550$ & $\mathbf{0.0483}$ \\
        \bottomrule
    \end{tabular}
    \vspace{-8pt}
\end{wraptable}

Adding $z$ raises PSNR from $25.92$ to $27.27$\,dB and SSIM from $0.885$ to
$0.902$, while reducing LPIPS by $12.2\%$, from $0.0550$ to $0.0483$
(Table~\ref{tab:world_modeling_ability}). The agreement across distortion,
structural, and perceptual metrics makes a metric-specific explanation
unlikely. Moreover, $z$ is computed from the same visual and language context
already available to the video-only baseline, and it supplies no new observation.
The gain therefore indicates that the predictive encoder reorganizes the
context into a form that the generator can use more effectively.

Together, the two ablations validate opposite directions of the coupling.
Generator states provide a stronger target for learning $z$, while $z$ remains
useful to the generator rather than becoming a policy-specific summary. This
reciprocity is the main evidence for treating $z$ as a shared predictive
interface.

\subsection{What Enfold Encodes}
\label{sec:representation_analysis}

Task performance establishes the utility of generator-state supervision, but does not reveal what the learned representation encodes. Since generator states contain both transition-relevant structure and stochastic variation, we ask whether Enfold merely reproduces this nuisance or learns a genuinely future-oriented representation beyond static visual features.

\begin{wrapfigure}{r}{0.52\textwidth}
    \centering
    \vspace{-10pt}
    \includegraphics[width=\linewidth]
    {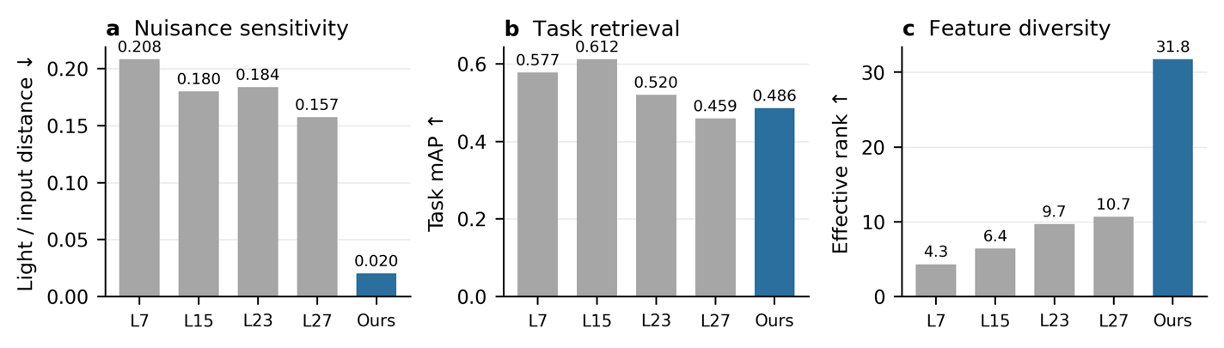}
    \caption{Comparison between Cosmos internal states ($t=600$) and the Enfold
    predictive encoder on a separate $100$-clip set. Lower lighting sensitivity
    is preferred. Task mAP and effective rank report task selectivity and feature
    diversity, respectively.}
    \label{fig:representation_stability}
    \vspace{-5pt}
\end{wrapfigure}

\paragraph{From stochastic teacher features to a stable predictive
representation.}
Raw generator states vary with both scene content and generation noise
(Section~\ref{sec:predictive_targets}). We compare the Enfold predictive encoder with
representative Cosmos layers under controlled lighting perturbations. Nuisance
sensitivity is the resulting feature displacement normalized by the variation
across input videos.

The predictive representation reduces this ratio from $0.157$--$0.208$ for the
generator layers to $0.020$, a $7.9$--$10.4\times$ reduction
(Figure~\ref{fig:representation_stability}). On this $100$-clip set, its
effective rank is $31.8$, compared with at most $10.7$ for the generator
features, while
retaining a task-retrieval mAP of $0.486$. Lower sensitivity alone could be
explained by feature collapse, but the higher effective rank argues against that
explanation. The student is therefore not a compressed copy of any single
teacher layer. Its behavior is consistent with predictive learning acting as a
conditional filter: variation that cannot be inferred reliably from the present
is suppressed, while shared structure is reorganized into a more diverse
representation.

\paragraph{Predicting future change.}
We next probe whether the difference from frozen DINO is specifically
future-oriented. Equal-capacity MLP probes predict frozen-DINO features of a
future frame from either the Enfold representation or frozen-DINO features of
the current frame. We report global prediction and performance on the patches that change
most over the future horizon. The latter subset is selected using the observed future only for analysis and is not available at inference.

\begin{figure*}[t]
    \centering
    \includegraphics[width=1.0\textwidth]{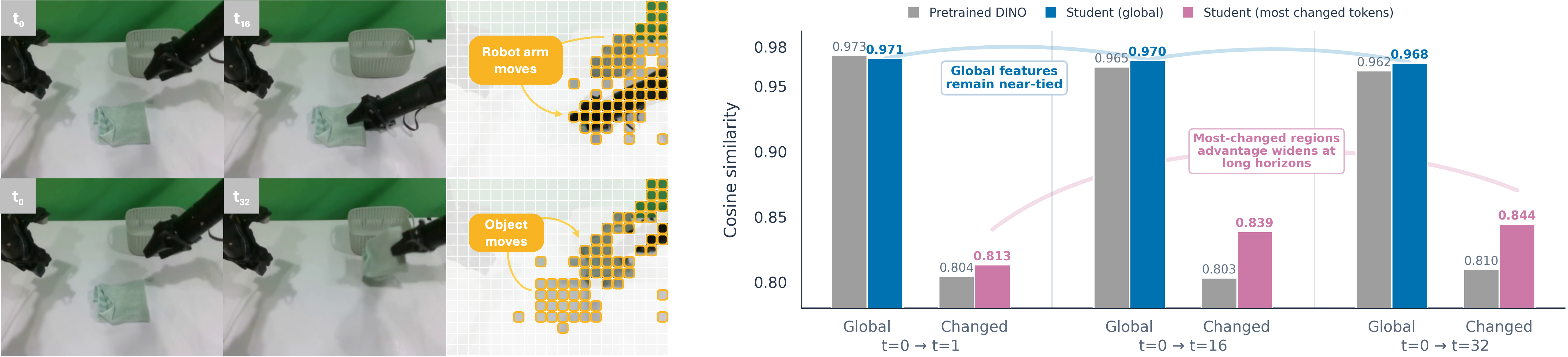}
    \caption{
    \textbf{Future-observation prediction probe and changed-token localization.} \textbf{Left:} Visualization
    of the selected most-changing tokens for two future horizons. The selected locations concentrate on the moving robot arms and the manipulated cloth, rather than the static tabletop or background. \textbf{Right:} Compared with frozen DINO, Enfold's advantage on the changed-token subset grows with future time and remains strong at long horizons.
    }
    \label{fig:student_future_probe}
\end{figure*}

The advantage is most diagnostic in regions that move
(Figure~\ref{fig:student_future_probe}). On the selected patches, the Enfold
representation reduces residual cosine error (1-cosine) by $4.6\%$, $18.1\%$, and $18.2\%$ at future time
$t=1,16,32$. These patches lie primarily on the arms and manipulated cloth.
With global pooling, Enfold is slightly worse at $t=1$ but reduces residual error
by $14.0\%$ and $15.2\%$ at $t=16$ and $t=32$. This short-horizon exception is
diagnostic. If Enfold were merely a uniformly stronger image descriptor, it
should also dominate on the nearly unchanged next frame. Instead, its advantage
grows with temporal distance and is largest on patches that actually change.
The evidence therefore localizes the benefit to future-dependent content rather
than generic visual quality.

\begin{figure*}[t]
    \centering
    \includegraphics[width=0.98\textwidth]{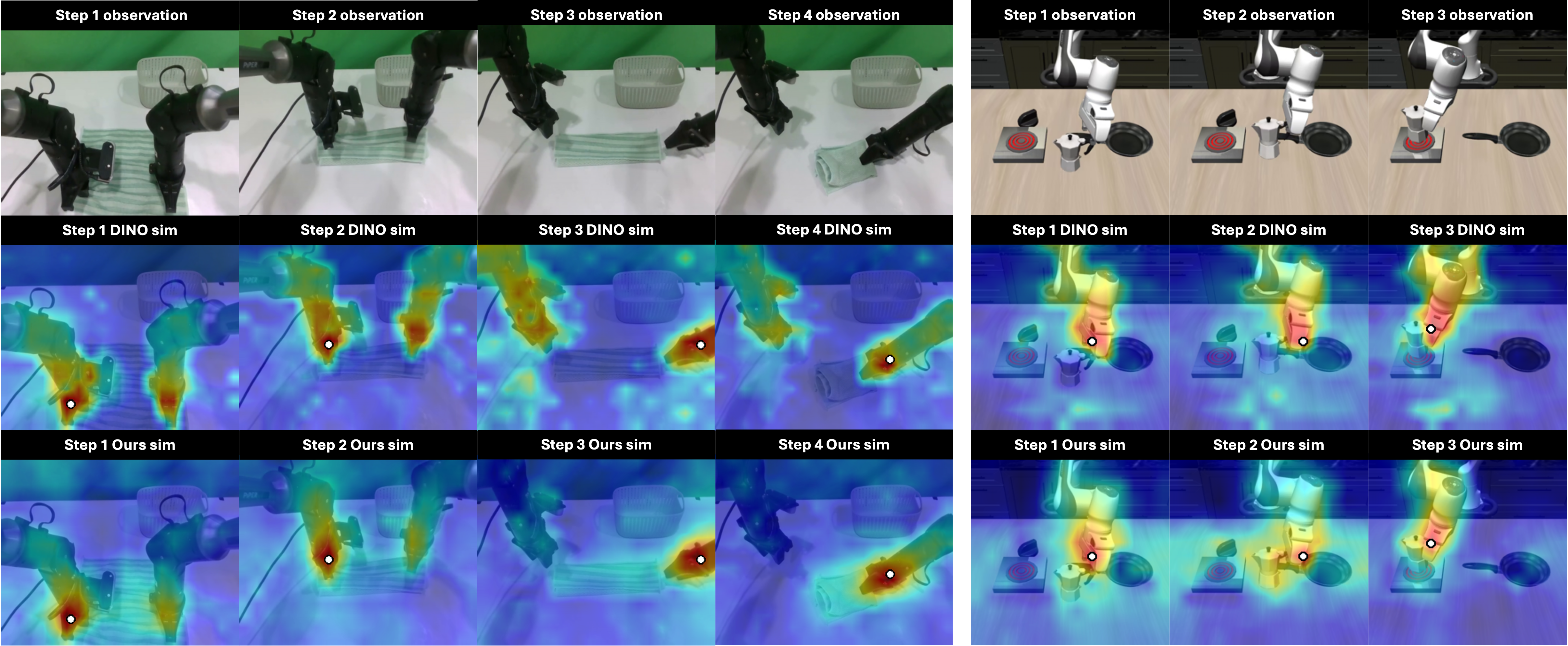}
    \caption{
    \textbf{Query-point similarity overlays.} The left four columns show a rollout of the real-world fold-towel task, while the right three columns show LIBERO trajectories. White dots mark manually selected query points on the robot gripper. The middle row shows cosine-similarity overlays from frozen DINO patch tokens, while the bottom row shows overlays from the Enfold predictive encoder. Whereas frozen DINO mainly responds to visually similar arm regions, Enfold also highlights the manipulated object and contact-relevant regions that subsequently change during manipulation.}
    \label{fig:student_point_similarity}
\end{figure*}

\paragraph{Interaction structure in the token geometry.}
The fitted probe establishes future predictability but not how it is organized
spatially. We therefore inspect the representation without a learned readout.
For each observation, a query point on the gripper is mapped to its patch token,
and we compute $\cos(q_t,k_{t,j})$ against all tokens in the same frame. Frozen
DINO and Enfold are evaluated independently at the same query location.

Frozen DINO mainly associates the query with visually similar parts of the arm
(Figure~\ref{fig:student_point_similarity}). In Enfold, high similarity extends
from the gripper to the manipulated object and the regions involved in contact.
The pattern appears before contact and persists while the object is moved, in
both real and simulated examples. The cross-object association cannot be
explained by local appearance similarity alone, and it is consistent with the token
geometry organizing an interaction relation between the agent and the object.
This visualization remains qualitative, but it complements the fitted probe:
generator-mediated supervision shifts the representation toward the parts of
the scene that participate in the future interaction.

\section{Discussion and Conclusion}

World generative models are commonly treated as simulators: their knowledge becomes useful by generating a future or by remaining active in the control loop. Enfold reveals another possibility. By predicting the generator's future-conditioned internal states from the current observation and instruction, it transfers the generator's organization of future change into a representation available from the present. This representation is not shaped by action gradients, yet remains useful for both future generation and control. Its coordinated adaptation of imagination and action under human intervention further suggests that it is recomputed from the altered scene rather than tied to a memorized trajectory.

Enfold therefore does more than remove video decoding from a world-action model. It changes the role of the generator from an action-time simulator to a training-time source of structured predictive supervision. The generator is still required during training, and the intervention evidence remains qualitative. Nevertheless, the results establish a broader principle: rich generative computation can be internalized during learning and reused efficiently for control, while explicit imagination is invoked only when needed.

\section*{Acknowledgments}
The authors appreciate the GPU resources provided by the National Supercomputing Center in Jinan. This work was supported in part by the CCF--Baidu Open Fund.

\bibliographystyle{unsrt}
\bibliography{sample}

\newpage
\appendix

\section{A Predictive-Projection Interpretation of G2R}
\label{app:predictive_projection}

G2R uses future-conditioned generator states as supervision even though the
future itself is unavailable at deployment. What, then, can a current-only
encoder learn from targets that also depend on the realized future and the
stochastic generative process? We give an idealized predictive-projection
interpretation. At a fixed stage of training, we hold the target-producing
computation---the generator and the detached map from the current context to
$\bar z$---fixed, and consider the population regression problem it induces.
The analysis characterizes the direction of a G2R update. It is not a convergence claim for the jointly evolving training system.

Let $X=(c,e)$ collect the current visual context and language instruction, and
let $Y\sim p(y\mid X)$ denote a realized future. Together with
$\epsilon\sim\mathcal{N}(0,I)$ and corruption timestep $t$, the
teacher-forced pass in Eq.~\ref{eq:teacher_target} defines a random target
$r_t^G=r_t^G(X,Y,\epsilon)$. Although this target is deterministic once
$(X,Y,\epsilon,t)$ are fixed, it remains random from the perspective of the
current context. Consider the squared-error counterpart of
Eq.~\ref{eq:g2r},
\begin{equation}
    \widetilde{\mathcal{L}}_{\mathrm{G2R}}(f)
    =
    \mathbb{E}
    \left[
        \left\|f(X,t)-r_t^G\right\|_2^2
    \right].
\end{equation}
For an unrestricted predictor and a fixed induced target distribution, the
population optimum is
\begin{equation}
    f^\star(X,t)
    =
    \mathbb{E}\!\left[r_t^G\mid X,t\right].
\end{equation}
This conditional expectation is the $L_2$ projection of the generator state
onto functions of the information available to the predictor. In particular,
writing
\begin{equation}
    r_t^G
    =
    \underbrace{\mathbb{E}[r_t^G\mid X,t]}_{r_t^{G,\mathrm{pred}}}
    +
    \underbrace{
        \left(r_t^G-\mathbb{E}[r_t^G\mid X,t]\right)
    }_{r_t^{G,\mathrm{res}}},
    \qquad
    \mathbb{E}[r_t^{G,\mathrm{res}}\mid X,t]=0,
\end{equation}
gives, for any predictor $f(X,t)$,
\begin{equation}
\begin{aligned}
    \mathbb{E}\!\left[\|r_t^G-f(X,t)\|_2^2\right]
    &=
    \mathbb{E}\!\left[\|r_t^{G,\mathrm{res}}\|_2^2\right]
    +
    \mathbb{E}\!\left[
        \|r_t^{G,\mathrm{pred}}-f(X,t)\|_2^2
    \right].
\end{aligned}
\label{eq:predictive_projection_decomposition}
\end{equation}
The residual term includes variation due to unresolved future realizations,
corruption noise, and any other target component that is not systematically
predictable from $(X,t)$. At the population level, reducing G2R cannot recover
this term. It instead rewards the component of the generator computation that
covaries predictably with the current scene and instruction. This does not rule
out finite-sample memorization, but it explains why G2R should not be viewed as
unconditional feature copying. It is also consistent with the substantially
lower nuisance sensitivity of the learned representation reported in
Section~\ref{sec:representation_analysis}.

The implemented objective only approximates this idealized projection.
$F_\omega\circ U_\phi$ is a finite-capacity function class, the target-producing
branch changes over training, and Eq.~\ref{eq:g2r} uses SmoothL1 rather than
squared error. Under SmoothL1, the population optimizer is a robust conditional
center instead of the conditional mean exactly. It approaches mean regression
in the quadratic regime while limiting the influence of large target
deviations. The projection view should therefore be read as an interpretation
of what information the objective makes learnable, rather than as an exact
closed-form description of the trained network.

This view also clarifies the timestep-conditioned and multi-level design. The
encoder produces a shared state $z=U_\phi(X)$ that is independent of $t$, while
$F_\omega(z,\gamma(t))$ predicts the target in the coordinate system induced
at timestep $t$. Thus, $z$ must support prediction of the family
\begin{equation}
    \left\{
        \mathbb{E}[r_t^G\mid X,t]
    \right\}_{t\in\mathcal{T}},
\end{equation}
and the prediction head accounts for how that structure is expressed along the
generative trajectory. Concatenating several generator depths extends this
family across abstraction levels, allowing the predictor to draw on
complementary structure without assuming that any single layer is universally
optimal.

Finally, the interpretation distinguishes predictive organization from
information gain. For a fixed trained encoder, $Z=U_\phi(X)$ is a deterministic
function of the deployment-time input, and therefore
\begin{equation}
    I(Y;Z\mid X)=0.
\end{equation}
Conditioning the generator on $Z$ cannot reveal sample-specific future
information absent from $X$. An R2G improvement instead indicates that $Z$
reorganizes predictive structure already present in the context into
coordinates that are more accessible to a finite-capacity generator. Detached
task heads ask the complementary question of whether the same organization is
actionable without allowing task gradients to define it.

Under these qualifications, Enfold can be viewed as learning an amortized,
robust approximation to
\begin{equation}
    \boxed{
    r_t^G
    =
    \mathbb{E}[r_t^G\mid c,e,t]
    +
    r_t^{G,\mathrm{res}}
    }.
\end{equation}
The conditional expectation is the predictable generative structure, whereas
the residual contains unresolved future and stochastic variation. G2R makes
the first component accessible from the present, while R2G and the detached
task readouts test whether the resulting representation remains useful for
generation and control.

\section{Experimental Details}
\label{app:experimental_details}

\subsection{Benchmarks and Model Configuration}

We use the four standard $10$-task LIBERO
suites~\cite{liu2023libero}. After removing no-operation transitions, we train
on $1712$ demonstrations and evaluate each task with $50$ trials. RoboTwin2.0
provides $50$ dual-arm tasks and structured domain randomization
axes~\cite{chen2025robotwin2}. Our training set contains $27500$
demonstrations, and evaluation uses the clean and randomized scene
configurations. The real-world dataset contains $400$ demonstrations from four
tabletop tasks collected with an AgileX dual-arm platform.

\paragraph{Backbones and feature routing.}
All three embodiments use DINOv3 ViT-H+/16 as the predictive visual
encoder~\cite{simeoni2025dinov3}. LIBERO supplies two views at
$224\times448$, the real-robot policy uses one frontal $256\times320$ view, and
RoboTwin2.0 uses front and bilateral wrist views at $384\times320$.
Cosmos-Reason1-7B language
features~\cite{nvidia2025cosmosreason1} are linearly projected and concatenated
into the DINO stream from block 8. The action head reads projected patch tokens
from the final DINO block. Features from blocks $\{7,15,23,31\}$ are instead
concatenated only for the predictive head and its multi-level supervision.
Detached features from Cosmos blocks $\{7,15,23,27\}$ provide the G2R
targets~\cite{nvidia2025cosmoswfm}, and the predicted representation is fed back to
the same generator through R2G.

\paragraph{Prediction and action heads.}
The predictive head has $10$ layers, maps
$5120\!\rightarrow\!1024\!\rightarrow\!8192$, and combines its conditions by
concatenation with 3D RoPE. The action head uses the same concatenative
conditioning with 1D RoPE. It contains $10$ layers for LIBERO and real-robot
control and $16$ layers for RoboTwin2.0.

\paragraph{Temporal processing, inference, and optimization.}
Each observation conditions a $32$-step action chunk and a corresponding video
sequence containing one context frame and $32$ future frames. Temporal
subsampling with stride 4 reduces the video sequence to $1+8$ frames. The causal
VAE further compresses it to $1+2$ latent frames. Actions are trained with
continuous flow matching from Gaussian noise and sampled with a $10$-step Euler
solver using a noise shift of $5.0$. The policy replans after executing $10$
actions on LIBERO and the real robot and $24$ actions on RoboTwin2.0. All models
use bf16, AdamW with a learning rate of $10^{-4}$, a cosine schedule, and unit
coefficients for the R2G and task losses. Enfold-Flash retains the same
representation interface and adds TensorRT operator
acceleration~\cite{nvidia2026tensorrt}. Internally trained baselines and
ablations use the same observations, language conditions, task data, and
evaluation protocols. Table~\ref{tab:training_configuration} summarizes the
dataset-specific settings.

\begin{table}[!t]
\centering
\caption{\textbf{Training and policy configuration.}}
\label{tab:training_configuration}
\small
\setlength{\tabcolsep}{4pt}
\resizebox{\linewidth}{!}{
\begin{tabular}{lccc}
\toprule
Configuration & LIBERO & Real Robot & RoboTwin2.0 \\
\midrule
\multicolumn{4}{l}{\textit{Encoder and feature prediction}} \\
Predictive encoder
    & DINOv3 ViT-H+/16
    & DINOv3 ViT-H+/16
    & DINOv3 ViT-H+/16 \\
Camera views
    & 2 (front, wrist)
    & 1 (front)
    & 3 (front, left wrist, right wrist) \\
Input resolution
    & $224\times448$
    & $256\times320$
    & $384\times320$ \\
Encoder positional encoding
    & 2D RoPE & 2D RoPE & 2D RoPE \\
Text--DINO concat start layer
    & 8 & 8 & 8 \\
DINO feature concat layers
    & $\{7,15,23,31\}$
    & $\{7,15,23,31\}$
    & $\{7,15,23,31\}$ \\
Generative-model feature layers
    & $\{7,15,23,27\}$
    & $\{7,15,23,27\}$
    & $\{7,15,23,27\}$ \\
Generative condition
    & Concat, detached
    & Concat, detached
    & Concat, detached \\
\midrule
\multicolumn{4}{l}{\textit{Prediction and action heads}} \\
Prediction-head depth
    & 10 & 10 & 10 \\
Prediction-head dimensions
    & $5120\!\rightarrow\!1024\!\rightarrow\!8192$
    & $5120\!\rightarrow\!1024\!\rightarrow\!8192$
    & $5120\!\rightarrow\!1024\!\rightarrow\!8192$ \\
Prediction-head condition injection
    & Concat & Concat & Concat \\
Prediction-head positional encoding
    & 3D RoPE & 3D RoPE & 3D RoPE \\
Action-head depth
    & 10 & 10 & 16 \\
Action-head hidden dimension
    & 1024 & 1024 & 1024 \\
Action-head condition injection
    & Concat & Concat & Concat \\
Action-head positional encoding
    & 1D RoPE & 1D RoPE & 1D RoPE \\
Diffusion-time embedding dimension
    & 256 & 256 & 256 \\
\midrule
\multicolumn{4}{l}{\textit{Temporal horizons}} \\
Action context / future horizon
    & $1/32$ steps
    & $1/32$ steps
    & $1/32$ steps \\
Raw-video context / future horizon
    & $1/32$ frames
    & $1/32$ frames
    & $1/32$ frames \\
Temporal video sampling stride
    & 4 & 4 & 4 \\
VAE-input context / future horizon
    & $1/8$ frames
    & $1/8$ frames
    & $1/8$ frames \\
VAE temporal compression ratio
    & 4 & 4 & 4 \\
Latent-video context / future horizon
    & $1/2$ latents
    & $1/2$ latents
    & $1/2$ latents \\
\midrule
\multicolumn{4}{l}{\textit{Action generation and execution}} \\
Action generation objective
    & Flow matching & Flow matching & Flow matching \\
Initial action distribution
    & $\mathcal{N}(0,I)$
    & $\mathcal{N}(0,I)$
    & $\mathcal{N}(0,I)$ \\
Training noise timesteps
    & 1000 & 1000 & 1000 \\
Train / inference noise shift
    & $5.0/5.0$ & $5.0/5.0$ & $5.0/5.0$ \\
Action sampling solver
    & Euler & Euler & Euler \\
Action sampling steps
    & 10 & 10 & 10 \\
Executed actions per replanning
    & 10 & 10 & 24 \\
Guidance scale
    & 1.0 & 1.0 & 1.0 \\
\midrule
\multicolumn{4}{l}{\textit{Training}} \\
Training hardware
    & 1 node / $8{\times}$ A100 40GB
    & 1 node / $8{\times}$ A100 40GB
    & 4 nodes / $32{\times}$ A100 40GB \\
Wall-clock training time
    & 23 h & 44 h & 108 h \\
Global batch size
    & 128 & 128 & 320 \\
Optimizer and learning rate
    & AdamW / $1{\times}10^{-4}$
    & AdamW / $1{\times}10^{-4}$
    & AdamW / $1{\times}10^{-4}$ \\
Learning-rate scheduler
    & Cosine & Cosine & Cosine \\
$\lambda_{\mathrm{R2G}},\lambda_{\mathrm{task}}$
    & $1,1$ & $1,1$ & $1,1$ \\
Training epochs
    & 10 & 10 & 5 \\
Training steps
    & 21,700 & 40,800 & 94,925 \\
Numerical precision
    & bf16 & bf16 & bf16 \\
\bottomrule
\end{tabular}
}
\end{table}

\subsection{Control Evaluation Protocols}

\begin{table*}[t]
\centering
\caption{\textbf{Per-task success rates on RoboTwin2.0 under clean and
randomized evaluation settings.} The best value(s) in each row and setting are
shown in \best{bold}. The second-best distinct value(s) are
\second{underlined}.}
\label{tab:robotwin_per_task}
\scriptsize
\setlength{\tabcolsep}{2.8pt}
\resizebox{\textwidth}{!}{
\begin{tabular}{l*{6}{cc}}
\toprule
& \multicolumn{2}{c}{Enfold}
& \multicolumn{2}{c}{Enfold-Flash}
& \multicolumn{2}{c}{Fast-WAM}
& \multicolumn{2}{c}{LingBot-VA}
& \multicolumn{2}{c}{$\pi_{0.5}$}
& \multicolumn{2}{c}{Motus} \\
\cmidrule(lr){2-3}
\cmidrule(lr){4-5}
\cmidrule(lr){6-7}
\cmidrule(lr){8-9}
\cmidrule(lr){10-11}
\cmidrule(lr){12-13}
Task
& Clean & Rand.
& Clean & Rand.
& Clean & Rand.
& Clean & Rand.
& Clean & Rand.
& Clean & Rand. \\
\midrule
Adjust Bottle
& \best{100\%} & \best{100\%} & \best{100\%} & \best{100\%} & \best{100\%} & \best{100\%} & \second{90\%} & 94\% & \best{100\%} & \second{99\%} & 89\% & 93\% \\
Beat Block Hammer
& 97\% & \best{98\%} & \second{98\%} & 94\% & \best{99\%} & \second{97\%} & 96\% & \best{98\%} & 96\% & 93\% & 95\% & 88\% \\
Blocks Ranking RGB
& \second{99\%} & 95\% & \second{99\%} & 94\% & \best{100\%} & \best{100\%} & \second{99\%} & \second{98\%} & 92\% & 85\% & \second{99\%} & 97\% \\
Blocks Ranking Size
& \second{91\%} & 95\% & \second{91\%} & 91\% & \best{94\%} & \best{98\%} & \best{94\%} & \second{96\%} & 49\% & 26\% & 75\% & 63\% \\
Click Alarmclock
& 97\% & \best{100\%} & \best{100\%} & \best{100\%} & \best{100\%} & \best{100\%} & \second{99\%} & \best{100\%} & 98\% & \second{89\%} & \best{100\%} & \best{100\%} \\
Click Bell
& \best{100\%} & \best{100\%} & \best{100\%} & \best{100\%} & \best{100\%} & \best{100\%} & \best{100\%} & \best{100\%} & \second{99\%} & \second{66\%} & \best{100\%} & \best{100\%} \\
Dump Bin Bigbin
& \best{99\%} & 93\% & \second{97\%} & 95\% & \second{97\%} & \second{96\%} & 89\% & \second{96\%} & 92\% & \best{97\%} & 95\% & 91\% \\
Grab Roller
& \best{100\%} & \best{100\%} & \best{100\%} & \best{100\%} & \best{100\%} & \best{100\%} & \best{100\%} & \best{100\%} & \best{100\%} & \best{100\%} & \best{100\%} & \best{100\%} \\
Handover Block
& 93\% & 73\% & 86\% & \second{78\%} & \second{95\%} & \best{81\%} & \best{99\%} & \second{78\%} & 66\% & 57\% & 86\% & 73\% \\
Handover Mic
& \second{99\%} & \best{100\%} & \best{100\%} & \best{100\%} & \second{99\%} & \best{100\%} & 94\% & 96\% & 98\% & \second{97\%} & 78\% & 63\% \\
Hanging Mug
& 48\% & \second{45\%} & \second{51\%} & 44\% & \best{58\%} & \best{62\%} & 40\% & 28\% & 18\% & 17\% & 38\% & 38\% \\
Lift Pot
& 97\% & \best{100\%} & \second{99\%} & \best{100\%} & \best{100\%} & \best{100\%} & \best{100\%} & \second{99\%} & 96\% & 85\% & 96\% & \second{99\%} \\
Move Can Pot
& \best{95\%} & 93\% & 93\% & \second{96\%} & 90\% & 88\% & \second{94\%} & \best{97\%} & 51\% & 55\% & 34\% & 74\% \\
Move Pillbottle Pad
& 98\% & \second{98\%} & 96\% & 97\% & \best{100\%} & \best{99\%} & \second{99\%} & \best{99\%} & 84\% & 61\% & 93\% & 96\% \\
Move Playingcard Away
& \second{98\%} & \best{100\%} & \second{98\%} & \second{99\%} & \best{100\%} & \best{100\%} & \best{100\%} & \second{99\%} & 96\% & 84\% & \best{100\%} & 96\% \\
Move Stapler Pad
& 72\% & 71\% & 62\% & 74\% & 77\% & 64\% & \best{91\%} & \second{79\%} & 56\% & 42\% & \second{83\%} & \best{85\%} \\
Open Laptop
& \second{97\%} & \second{99\%} & \best{98\%} & \second{99\%} & \best{98\%} & \best{100\%} & 92\% & 94\% & 90\% & 96\% & 95\% & 91\% \\
Open Microwave
& 83\% & \second{94\%} & \second{89\%} & \best{97\%} & 62\% & 45\% & 82\% & 86\% & 34\% & 77\% & \best{95\%} & 91\% \\
Pick Diverse Bottles
& 80\% & 81\% & 78\% & \second{85\%} & 80\% & \second{85\%} & \second{89\%} & 82\% & 81\% & 71\% & \best{90\%} & \best{91\%} \\
Pick Dual Bottles
& 88\% & 93\% & 89\% & 87\% & \best{100\%} & \second{96\%} & \best{100\%} & \best{99\%} & 93\% & 63\% & \second{96\%} & 90\% \\
Place A2B Left
& \second{95\%} & \best{95\%} & 94\% & \best{95\%} & \second{95\%} & \second{93\%} & \best{97\%} & \second{93\%} & 87\% & 82\% & 88\% & 79\% \\
Place A2B Right
& \second{93\%} & 91\% & \second{93\%} & 92\% & \second{93\%} & \best{99\%} & \best{97\%} & \second{95\%} & 87\% & 84\% & 91\% & 87\% \\
Place Bread Basket
& 93\% & \best{96\%} & \second{96\%} & \second{95\%} & 91\% & 93\% & \best{97\%} & \second{95\%} & 77\% & 64\% & 91\% & 94\% \\
Place Bread Skillet
& \second{93\%} & 87\% & 91\% & \second{91\%} & 90\% & \best{93\%} & \best{95\%} & 90\% & 85\% & 66\% & 86\% & 83\% \\
Place Burger Fries
& \second{97\%} & \best{100\%} & \best{98\%} & \second{99\%} & 96\% & \second{99\%} & \second{97\%} & 95\% & 94\% & 87\% & \best{98\%} & 98\% \\
Place Can Basket
& \second{75\%} & 63\% & \second{75\%} & 62\% & 71\% & 69\% & \best{81\%} & \best{84\%} & 62\% & 62\% & \best{81\%} & \second{76\%} \\
Place Cans Plasticbox
& \second{99\%} & \best{99\%} & \second{99\%} & \second{97\%} & \second{99\%} & 96\% & \best{100\%} & \best{99\%} & 94\% & 84\% & 98\% & 94\% \\
Place Container Plate
& \best{100\%} & 97\% & \best{100\%} & 98\% & 96\% & \best{100\%} & \second{99\%} & 97\% & \second{99\%} & 95\% & 98\% & \second{99\%} \\
Place Dual Shoes
& 86\% & \best{90\%} & 92\% & 88\% & \best{94\%} & 88\% & \best{94\%} & \second{89\%} & 75\% & 75\% & \second{93\%} & 87\% \\
Place Empty Cup
& \best{100\%} & \best{100\%} & \best{100\%} & \best{100\%} & \best{100\%} & \best{100\%} & \best{100\%} & \best{100\%} & \best{100\%} & \second{99\%} & \second{99\%} & 98\% \\
Place Fan
& 93\% & 90\% & 93\% & \second{94\%} & \second{96\%} & \best{96\%} & \best{99\%} & 93\% & 87\% & 85\% & 91\% & 87\% \\
Place Mouse Pad
& 90\% & 94\% & \best{94\%} & \second{95\%} & 83\% & 89\% & \second{93\%} & \best{96\%} & 60\% & 39\% & 66\% & 68\% \\
Place Object Basket
& 81\% & 80\% & 84\% & 82\% & \second{89\%} & \best{88\%} & \best{91\%} & \best{88\%} & 80\% & 76\% & 81\% & \second{87\%} \\
Place Object Scale
& 92\% & \second{96\%} & \second{94\%} & \second{96\%} & 90\% & \best{97\%} & \best{96\%} & 95\% & 86\% & 80\% & 88\% & 85\% \\
Place Object Stand
& 95\% & 95\% & 95\% & \best{97\%} & 90\% & 94\% & \best{99\%} & \second{96\%} & 91\% & 85\% & \second{98\%} & \best{97\%} \\
Place Phone Stand
& \best{98\%} & \second{98\%} & 95\% & \best{99\%} & \second{97\%} & \best{99\%} & \second{97\%} & 97\% & 81\% & 81\% & 87\% & 86\% \\
Place Shoe
& 95\% & 97\% & 96\% & 95\% & 96\% & \best{99\%} & \second{98\%} & \second{98\%} & 92\% & 93\% & \best{99\%} & 97\% \\
Press Stapler
& \second{99\%} & \best{99\%} & \best{100\%} & \best{99\%} & 90\% & 97\% & 85\% & 82\% & 87\% & 83\% & 93\% & \second{98\%} \\
Put Bottles Dustbin
& \second{93\%} & 89\% & 89\% & \best{93\%} & \best{95\%} & 90\% & 87\% & \second{91\%} & 84\% & 79\% & 81\% & 79\% \\
Put Object Cabinet
& 83\% & \second{89\%} & 82\% & \best{90\%} & \best{94\%} & \second{89\%} & 85\% & 87\% & 80\% & 79\% & \second{88\%} & 71\% \\
Rotate QRcode
& 83\% & \second{89\%} & 88\% & 86\% & \second{93\%} & \second{89\%} & \best{96\%} & \best{91\%} & 89\% & 87\% & 89\% & 73\% \\
Scan Object
& 86\% & \best{93\%} & 84\% & 88\% & \second{89\%} & \second{92\%} & \best{96\%} & 91\% & 72\% & 65\% & 67\% & 66\% \\
Shake Bottle
& \best{100\%} & \best{100\%} & \best{100\%} & \best{100\%} & \best{100\%} & \best{100\%} & \best{100\%} & \second{97\%} & \second{99\%} & \second{97\%} & \best{100\%} & \second{97\%} \\
Shake Bottle Horizontally
& \best{100\%} & \best{100\%} & \best{100\%} & \best{100\%} & \best{100\%} & \best{100\%} & \best{100\%} & \second{99\%} & \second{99\%} & \second{99\%} & \best{100\%} & 98\% \\
Stack Blocks Three
& 96\% & 96\% & \second{98\%} & \second{97\%} & 95\% & \second{97\%} & \best{99\%} & \best{98\%} & 91\% & 76\% & 91\% & 95\% \\
Stack Blocks Two
& \best{100\%} & \best{100\%} & \best{100\%} & \best{100\%} & \best{100\%} & \best{100\%} & \best{100\%} & \second{98\%} & \second{97\%} & \best{100\%} & \best{100\%} & \second{98\%} \\
Stack Bowls Three
& \second{85\%} & 84\% & 82\% & \best{92\%} & 80\% & 81\% & \best{86\%} & 83\% & 77\% & 71\% & 79\% & \second{87\%} \\
Stack Bowls Two
& \second{97\%} & \best{99\%} & 96\% & 95\% & 92\% & \second{98\%} & 94\% & \second{98\%} & 95\% & 96\% & \best{98\%} & \second{98\%} \\
Stamp Seal
& 81\% & \second{94\%} & 86\% & 93\% & 90\% & \second{94\%} & \best{96\%} & \best{97\%} & 79\% & 55\% & \second{93\%} & 92\% \\
Turn Switch
& 71\% & \second{69\%} & \second{80\%} & 66\% & 61\% & 59\% & 44\% & 45\% & 62\% & 54\% & \best{84\%} & \best{78\%} \\
\midrule
Average
& 91.60\% & \second{91.94\%}
& \second{91.96\%} & \best{92.08\%}
& 91.88\% & 91.78\%
& \best{92.90\%} & 91.50\%
& 82.74\% & 76.76\%
& 88.66\% & 87.02\% \\
\bottomrule
\end{tabular}
}
\end{table*}

\paragraph{Real robot.}
We deploy on a tabletop AgileX dual-arm platform observed by one fixed frontal
RGB camera. Images are resized to $256\times320$. At each replanning step,
Enfold encodes the latest observation and language instruction, predicts a
$32$-step action chunk, and executes the first $10$ actions before acquiring a
new observation. The dataset contains $400$ teleoperated demonstrations, with
$100$ demonstrations for each of the four tasks.

\begin{table*}[t]
\centering
\caption{\textbf{Ordered task stages used for real-robot scoring.} Completing
stages 1--3 yields scores 1--3, respectively. Score 0 denotes no completed
stage.}
\label{tab:real_robot_protocol}
\small
\setlength{\tabcolsep}{4.5pt}
\begin{tabularx}{\textwidth}{@{}lXXX@{}}
\toprule
Task & Stage 1 & Stage 2 & Stage 3 (full completion) \\
\midrule
Fold Towel
& Form the first bimanual fold
& Reach the prescribed folded shape
& Complete the final placement of the folded towel \\
Organize Desktop
& Clear the first designated object group
& Place the remaining intermediate group
& Move all required objects to their target locations \\
Spoon Powder
& Acquire and align the spoon
& Scoop and transfer the powder
& Complete the transfer and return the tool \\
Store Plate
& Securely grasp the plate
& Reorient it for storage
& Place it stably in the target rack \\
\bottomrule
\end{tabularx}
\end{table*}

Following a staged-completion evaluation
protocol~\cite{cai2026ahawamasynchronoushorizonadaptiveworldactionmodeling}, we
run $N=30$ independent rollouts for each method, task, and evaluation setting.
If rollout $i$ completes $s_i\in\{0,1,2,3\}$ ordered stages, its normalized
completion score is
\begin{equation}
\operatorname{Score}
=
\frac{100}{3N}\sum_{i=1}^{N}s_i.
\label{eq:real_robot_score}
\end{equation}
A score of 3 is full task success. Lower scores retain partial progress. At a
timeout or safety-triggered termination, the score is fixed at the last
completed stage, and no rollout is discarded.

ID trials sample object poses and workspace configurations from the ranges used during data collection. OOD trials keep the instruction and completion criteria fixed while changing one factor whenever possible. Configuration shifts alter the towel geometry, object appearance, or the workspace background. All methods use the same initial-state sampler, trial allocation, camera observations, and termination limits within each setting.

\begin{figure*}[t]
    \centering
    \includegraphics[width=1.0\textwidth]{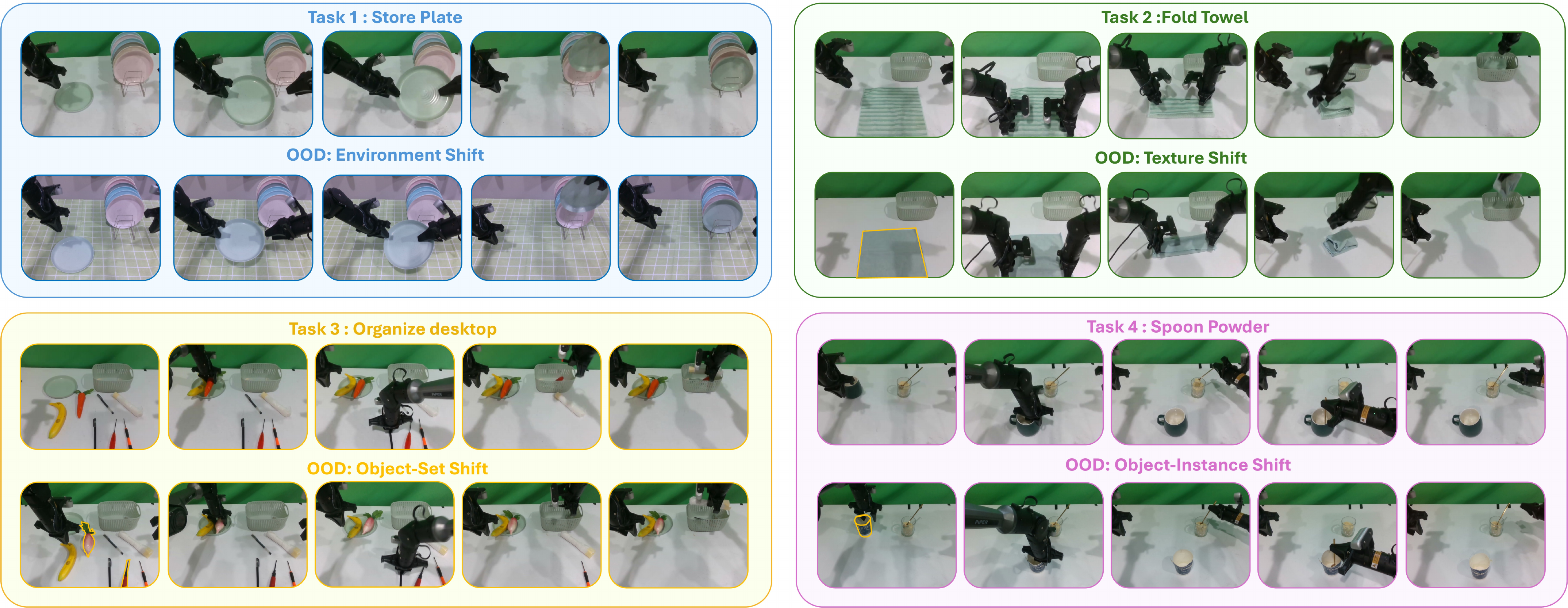}
    \caption{
    \textbf{Real-robot rollouts under in-distribution and OOD conditions.}
    Each panel shows representative Enfold executions for one task, with an in-distribution rollout on top and a matched OOD rollout below. The OOD setting changes the environment for \textit{Store Plate}, towel material for \textit{Fold Towel}, available object set for \textit{Organize Desktop}, and object instance for \textit{Spoon Powder}, while keeping the instruction and task-completion criterion fixed.}
    \label{fig:real_robot_task_ood}
\end{figure*}

\subsection{Future-Video Evaluation}

The video-only Cosmos baseline and Enfold are trained for the same $21.7$K
optimization steps and evaluated on $100$ fixed LIBERO demonstration clips.
Given the first frame and language instruction, each model predicts the next
$32$ frames with $35$ denoising steps. PSNR, SSIM, and LPIPS are computed only
over the predicted frames.

\subsection{Representation Analyses}

\paragraph{Stability.}
We compare the final Enfold predictive encoder with Cosmos layers L7, L15, L23,
and L27 at $t=600$ on $100$ clips and two noise seeds. Text conditioning is
disabled. Lighting sensitivity is the feature displacement under the controlled
illumination perturbation normalized by the mean displacement across different
input clips. For diversity, each model contributes one pooled feature per clip.
the two seed-specific Cosmos features are averaged before centering. Thus, for
model $M$, effective rank is evaluated on
$X^{M}_{100}\in\mathbb{R}^{100\times d_M}$ using
Equation~\ref{eq:effective_rank}, and is bounded by $99$. Task/scene retrieval
uses the same $100$ clip-level features. This evaluation is separate from the
$20$-video generator-state probe in Appendix~\ref{app:generator_probe}.

\paragraph{Future-feature probe.}
The probe uses $100$ real-world episodes of the fold-towel task: $80$ for
training and $20$ for validation. Equal-capacity MLPs predict the frozen-DINO feature
$z_t^{D}$ from either the Enfold representation $z_0^S$ or the current-frame DINO
feature $z_0^D$. For $X\in\{S,D\}$, we report
\[
s_t^X=\cos\!\left(f_t(z_0^X),z_t^D\right),
\qquad t\in\{1,16,32\},
\]
averaged over three probe seeds. The changed-region diagnostic uses the $k=64$
patches with the largest
future DINO displacement, $1-\cos(z_{0,j}^D,z_{t,j}^D)$, and both input
representations are evaluated at the same patch locations.

\paragraph{Token-similarity visualization.}
For each observation, we manually select a query point on the gripper and map
it to the corresponding patch token. We then compute its cosine similarity to
every token in the frame, independently for frozen DINO and the
instruction-conditioned Enfold predictive encoder.

\subsection{Human-Intervention Visualization}

The two episodes in Figure~\ref{fig:perturbation_recovery} begin from the
standard task distribution. At $t_2$, a human relocates the plate or displaces
the cloth without changing the language instruction. The next observation is
processed at the normal control frequency to obtain an updated predictive
representation. The middle panel is decoded from that representation only for
visualization. The physical rollout continues using actions predicted directly
from the representation. These examples are qualitative and are not used to
estimate a recovery rate. No matched frozen-representation or no-intervention
rollout is reported.

\subsection{Additional Qualitative Results}

\begin{figure*}[t]
    \centering
    \includegraphics[width=\textwidth]{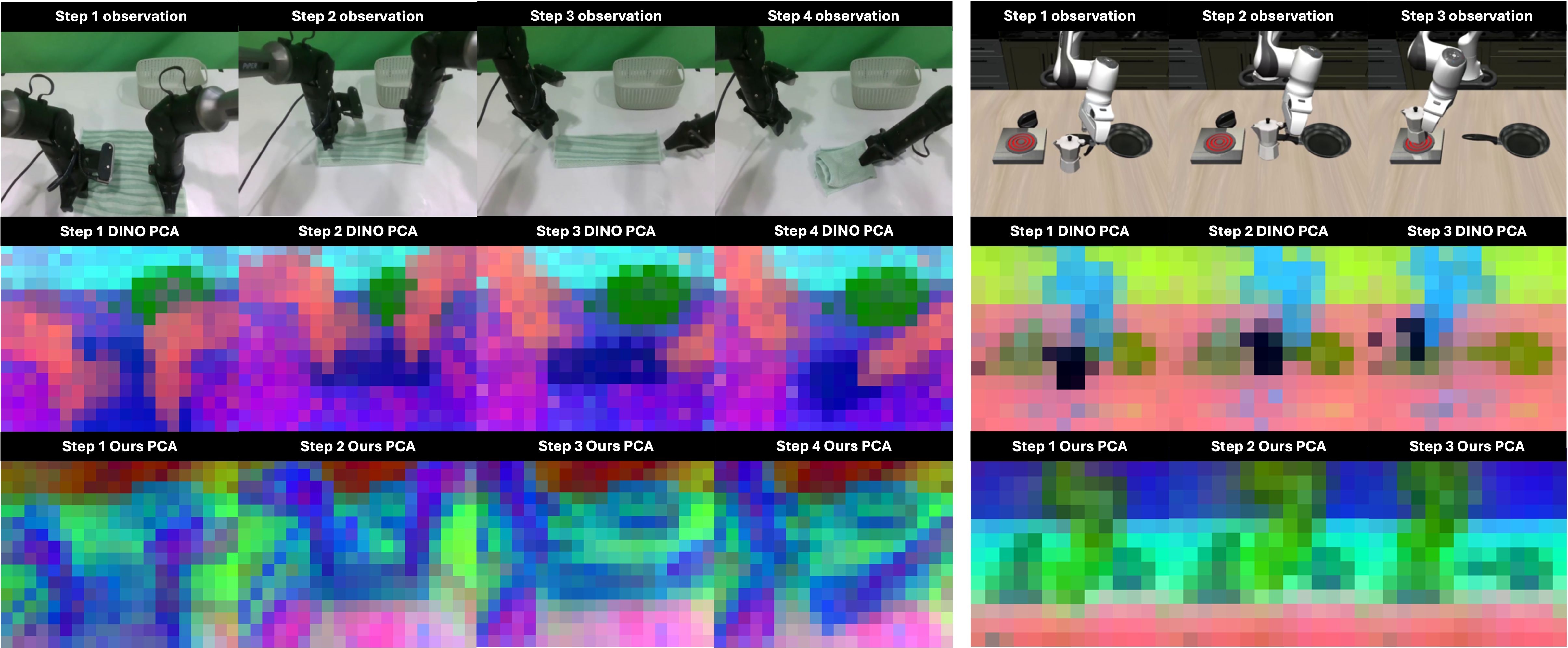}
    \caption{
    \textbf{PCA visualization of frozen DINO and Enfold token geometry.}
    The top row shows rollout observations. The middle and bottom rows visualize the corresponding spatial patch tokens from frozen DINO and the Enfold predictive encoder, respectively, by mapping their first three principal components to RGB. Frozen DINO separates the robot arm and task objects into visually distinct feature regions. Enfold makes the gripper and the object involved in the current manipulation more similar in the projection, forming a relational structure around the interaction.
    }
    \label{fig:token_pca_geometry}
\end{figure*}

\paragraph{PCA Visualization of Token Geometry}
Figure~\ref{fig:token_pca_geometry} provides a global view of the token
organization underlying the query-point similarity analysis in
Figure~\ref{fig:student_point_similarity}. For each observation, we project the spatial patch tokens to three principal components and map them to RGB.

Frozen DINO produces relatively sharp changes in the PCA map at visual object boundaries. The robot arm, manipulated object, tabletop, and background occupy distinct feature regions, consistent with a representation organized primarily by appearance and object identity. After generator-mediated training, Enfold changes this geometry near the manipulation. Across both the real-world fold-towel rollout and the LIBERO trajectory, the gripper and the object being operated on form more similar and spatially connected PCA regions, despite remaining visually distinct in the input image. Tokens outside the active interaction remain separated.

This complements Figure~\ref{fig:student_point_similarity}. The PCA maps show the broader reorganization of the token field, whereas the gripper-query cosine-similarity overlays directly measure the increased association between the gripper, manipulated object, and contact-relevant regions. Together, the two visualizations indicate that Enfold does not simply improve object discrimination. It reorganizes the representation around the interaction that determines the future manipulation.

\begin{figure*}[t]
    \centering
    \includegraphics[width=\textwidth]{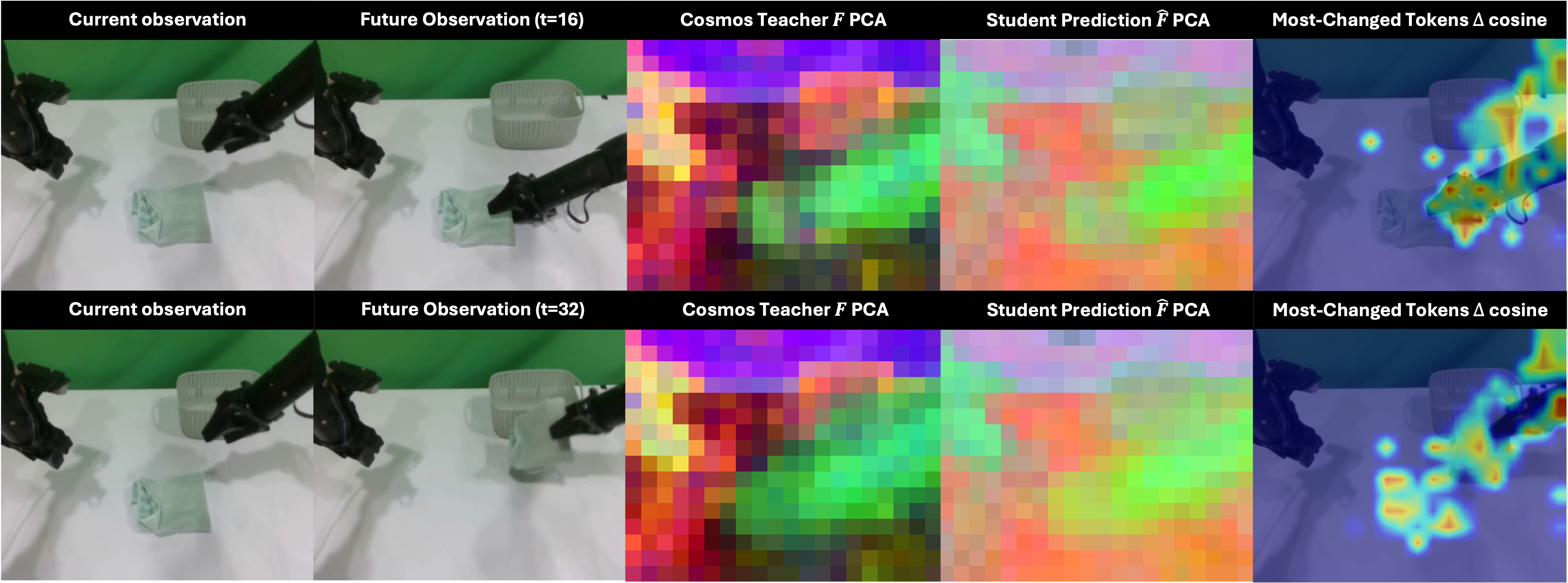}
    \caption{
    \textbf{Future alignment with teacher representations and prediction at most-changed tokens.}
    Rows show future horizons $t=16$ and $t=32$. From left to right: current and future observations, Cosmos teacher-target PCA, prediction-head output PCA, and the future-prediction gain $\Delta s_h=s_h^S-s_h^D$ on most-changed tokens.}
    \label{fig:teacher_feature_recovery}
\end{figure*}

\paragraph{Teacher-feature reconstruction and localization of future-prediction gains.}
The future-feature probe in Figure~\ref{fig:student_future_probe} uses fitted readouts to test whether the Enfold representation predicts a future frozen-DINO feature. Figure~\ref{fig:teacher_feature_recovery} instead visualizes the training-aligned G2R prediction directly. Given the current Enfold representation $z$, the prediction head produces
$\hat r^G_{\tau}=F_{\omega}(z,\gamma(\tau))$ to match the Cosmos teacher target $r^G_{\tau}$. At both displayed future horizons, the predicted features retain the coarse spatial organization of the teacher target, including the regions around the moving gripper, cloth, and their interaction.

The final column localizes the future-feature prediction result in
Figure~\ref{fig:student_future_probe}. It visualizes $\Delta s_h=s_h^S-s_h^D$ on the most-changed tokens, where $s_h^S$ and $s_h^D$ are the future-feature cosine scores obtained from Enfold and frozen DINO,
respectively. Most selected tokens have positive values: a readout from the Enfold representation predicts their future DINO features more accurately than a readout from the current frozen-DINO representation. The positive regions concentrate on the moving arms, manipulated cloth, and contact-relevant areas. This spatial pattern is consistent with the trend in Figure~\ref{fig:student_future_probe}: Enfold's advantage grows at longer horizons and is largest where the scene changes, rather than on static appearance-dominated regions.

\begin{figure*}[t]
    \centering
    \includegraphics[width=1.0\textwidth]{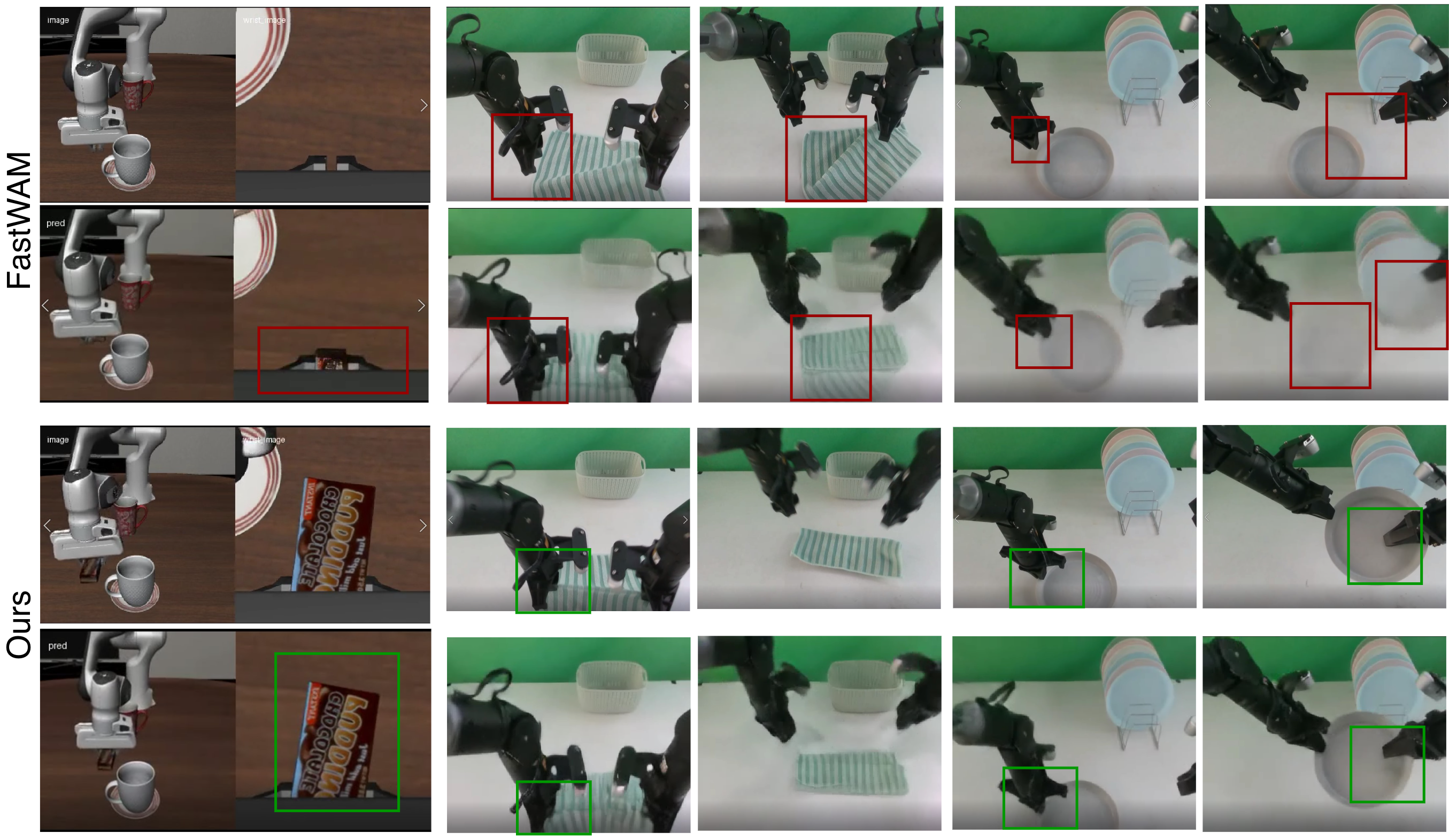}
    \caption{
    \textbf{Qualitative comparison of future-video predictions.}
    The left group shows a LIBERO task. The center and right groups show the real-robot \textit{Fold Towel} and \textit{Store Plate} tasks, respectively. The top two rows show FastWAM and the bottom two rows show Enfold. Within each method, the upper row contains rollout observations and the lower row contains predicted future frames. Red boxes mark mismatches
    in FastWAM predictions, including false object-in-gripper states in LIBERO and \textit{Fold Towel}, and a false plate grasp that leads to an empty-hand handover in \textit{Store Plate}. Green boxes mark the corresponding regions where Enfold predictions remain consistent with the observed future state.}
    \label{fig:fastwam_enfold_future_comparison}
\end{figure*}

\paragraph{Qualitative comparison of future-video artifacts.}
Figure~\ref{fig:fastwam_enfold_future_comparison} compares predicted future videos with the observations recorded during execution. The top two rows show FastWAM, and the bottom two rows show Enfold. For each method, the upper row is the executed rollout and the lower row is the predicted future video. The comparison focuses on object presence and gripper-object contact, since an
incorrect prediction of either can lead the policy to act on a state that has not actually occurred.

In the LIBERO example, FastWAM's wrist-view prediction places the object between the gripper fingers even though the object is still outside the gripper in the real rollout. The policy consequently continues to issue grasping motions, but cannot pick up the object. In \textit{Fold Towel}, FastWAM predicts that the left arm has already grasped the towel before the grasp is established. It then proceeds as if the towel were secured, while the ungrasped towel cannot be folded. In \textit{Store Plate}, FastWAM similarly
predicts that the left arm has lifted the plate although it has not. Later predicted frames still show a plate in the hand and lead to a handover motion, but the gripper is empty. In all three cases, Enfold's predicted frames agree with the observed object and contact states, avoiding these false grasp and handover states during execution.

% \begin{figure*}[t]
%     \centering
%     \placeholderbox{4.0cm}{
%     \textbf{Additional qualitative-results placeholder}\\[3pt]
%     Replace with a grid containing randomized RoboTwin2.0 examples,
%     real-robot successes, and representative failure cases.}
%     \caption{\textbf{Additional qualitative results and failure cases.} For
%     every example, show the input context, instruction, predicted or executed
%     trajectory, and final outcome. Failure cases should distinguish perception,
%     prediction, and control errors.}
%     \label{fig:additional_qualitative_results}
% \end{figure*}

\section{Generator-State Probe}
\label{app:generator_probe}

This appendix specifies the probe used in
Figures~\ref{fig:generator_target_hierarchy}
and~\ref{fig:generator_target_selectivity}. The probe is designed to compare
candidate supervision targets across generator depth and corruption level. It
does not evaluate complete video rollouts.

\subsection{Setup}

\paragraph{Data and controls.}
We probe a post-trained conditional video generator with $28$ Transformer
blocks (L0--L27). The dataset contains $20$ videos from four tabletop
manipulation tasks---\texttt{fold\_towel}, \texttt{organize\_desktop},
\texttt{spoon\_powder}, and \texttt{store\_plate}---with five episodes per
task. Each sample contains $33$ frames resized to $256\times320$. Short clips
are padded by repeating the final frame, and no temporal resampling is applied.

Every video is evaluated with two noise seeds and three illumination variants:
the original clip $x$, a dark variant $x^{1.45}$, and a bright variant
$x^{0.70}$, with pixel values clipped to $[0,1]$. All samples use the same
neutral prompt, \texttt{A robot arm operating on a tabletop.}, so task
retrieval cannot read the task identity directly from language. The complete
probe contains
$20\ \text{videos}\times3\ \text{illumination variants}\times2\ \text{seeds}
\times6\ \text{timesteps}=720$ teacher forward passes.

\paragraph{Corruption levels.}
Videos are encoded into VAE latents and evaluated at six corruption levels:
\[
z\in\mathbb{R}^{1\times16\times9\times32\times40},
\qquad
t\in\{0,300,600,800,900,999\}.
\]
At each level, we construct
\begin{equation}
z_{t,s}=(1-\sigma_t)z+\sigma_t\epsilon_{t,s},
\qquad
\sigma_t=t/1000,
\qquad
\epsilon_{t,s}\sim\mathcal{N}(0,I).
\label{eq:probe_corruption}
\end{equation}
The first latent temporal slice is replaced by the clean conditioning frame.
Thus, even at $t=999$, the generator receives one clean frame. Each probe point
uses a single teacher evaluation at a fixed timestep, rather than an iterative
denoising trajectory.

\subsection{Feature Extraction}

A teacher pass returns the output of every generator block,
\begin{equation}
H_{i,v,s}^{t,\ell}\in\mathbb{R}^{2880\times2048},
\end{equation}
where $i$, $v$, $s$, $t$, and $\ell$ index the video, illumination variant,
noise seed, corruption timestep, and generator block. All blocks from the same
pass share the same corrupted latent, timestep, and noise sample.

The probe uses the token-mean feature
\begin{equation}
f_{i,v,s}^{t,\ell}
=
\frac{1}{2880}\sum_{k=1}^{2880}H_{i,v,s,k}^{t,\ell}
\in\mathbb{R}^{2048}.
\label{eq:probe_pooling}
\end{equation}
This pooling is used only to characterize global feature geometry. Enfold's
G2R target retains spatio-temporal tokens. Feature displacement is measured
with cosine distance,
\begin{equation}
d(a,b)=1-\frac{a^\top b}{\|a\|_2\|b\|_2}.
\label{eq:probe_distance}
\end{equation}

\subsection{Metrics}

All quantities below are computed independently for each $(t,\ell)$ pair.
Let
$\bar f_i=\frac{1}{2}\sum_s f_{i,\mathrm{orig},s}$ denote the seed-averaged
feature of the original clip.

\paragraph{Input dispersion.}
The reference scale for representation change is the mean distance between
different videos,
\begin{equation}
D_{\mathrm{input}}
=
\frac{2}{N(N-1)}
\sum_{i<j}d(\bar f_i,\bar f_j),
\qquad N=20.
\label{eq:input_dispersion}
\end{equation}
Normalized sensitivities are interpreted only together with
$D_{\mathrm{input}}$, since a small ratio is not informative when the feature
itself is nearly constant.

\paragraph{Noise sensitivity.}
We measure the displacement caused by changing only the diffusion noise,
\begin{equation}
D_{\mathrm{seed}}
=
\frac{1}{N}\sum_i
d(f_{i,\mathrm{orig},0},f_{i,\mathrm{orig},1}),
\qquad
\rho_{\mathrm{seed}}=\frac{D_{\mathrm{seed}}}{D_{\mathrm{input}}}.
\label{eq:seed_sensitivity}
\end{equation}
Lower $\rho_{\mathrm{seed}}$ indicates that the representation is more stable
to the sampled generation noise. Numerical distances at $t=0$ are clipped to
zero.

\paragraph{Illumination sensitivity.}
Using seed-matched original/dark and original/bright pairs, we compute
\begin{equation}
D_{\mathrm{light}}
=
\frac{1}{2NS}
\sum_{i,s}
\left[
d(f_{i,\mathrm{orig},s},f_{i,\mathrm{dark},s})
+
d(f_{i,\mathrm{orig},s},f_{i,\mathrm{bright},s})
\right],
\qquad
\rho_{\mathrm{light}}=\frac{D_{\mathrm{light}}}{D_{\mathrm{input}}},
\label{eq:light_sensitivity}
\end{equation}
with $S=2$ seeds. Lower values indicate stronger invariance to the controlled
brightness perturbation.

\paragraph{Layout sensitivity.}
For two \texttt{organize\_desktop} episodes with different object
arrangements, denoted $a$ and $b$, we use
\begin{equation}
D_{\mathrm{layout}}
=
\frac{1}{S}\sum_s d(f_{a,\mathrm{orig},s},f_{b,\mathrm{orig},s}),
\qquad
\rho_{\mathrm{layout}}=\frac{D_{\mathrm{layout}}}{D_{\mathrm{input}}}.
\label{eq:layout_sensitivity}
\end{equation}
Unlike nuisance sensitivity, layout sensitivity is not minimized: object
arrangement can be relevant to manipulation. Figure~\ref{fig:generator_target_selectivity}
therefore compares $\rho_{\mathrm{layout}}$ with
$\rho_{\mathrm{light}}$ across depth.

\paragraph{Task retrieval.}
Each seed-averaged original feature retrieves the other $19$ videos by cosine
distance, and the four videos from the same task are positives. We report mean
average precision over the $20$ queries. Because task identity is correlated
with objects and scene layout, this metric is referred to as task/scene
retrieval rather than action recognition.

\paragraph{Effective rank in the generator-state probe.}
Let $X\in\mathbb{R}^{20\times2048}$ contain the centered features
$\{\bar f_i\}$. If $\{s_k\}$ are its singular values, we define
\begin{equation}
p_k=\frac{s_k^2}{\sum_j s_j^2},
\qquad
\operatorname{erank}(X)
=
\exp\left(-\sum_k p_k\log p_k\right).
\label{eq:effective_rank}
\end{equation}
Effective rank measures how broadly variation is distributed across feature
directions. Because $X$ is centered, the value in this $20$-video probe is at
most $19$. The $100$-clip comparison in
Figure~\ref{fig:representation_stability} uses the same estimator but a separate
matrix $X^{M}_{100}$, whose maximum rank is $99$. Raw values from the two sample
sets are therefore not compared. Within either set, effective rank is
interpreted jointly with $\rho_{\mathrm{seed}}$, since stochastic noise can
also increase rank.

\paragraph{Aggregation in the figures.}
Figures~\ref{fig:generator_target_hierarchy}(a,b) sweep task/scene retrieval
and $\rho_{\mathrm{seed}}$ over depth and timestep. Panel (c) shows their
trade-off at $t=600$, with marker size indicating effective rank.
Figure~\ref{fig:generator_target_selectivity} reports median responses across
$t\in\{300,600,800,900\}$ and uses the interquartile range over these four
timesteps. These statistics are descriptive and are not used for formal
significance testing.

\subsection{Scope and Limitations}

The probe is intentionally small and controlled. Its $20$ videos are sufficient
to expose depth- and timestep-dependent trends, but not to establish a precise
ranking of layers. Task retrieval is confounded with object and scene identity.
The illumination intervention changes global intensity and contrast but does
not model shadows, color temperature, or view-dependent reflectance, and the
layout response is estimated from one episode pair.

Token averaging suppresses local spatial structure, so a weak pooled response
does not imply that a layer lacks useful patch-level information. Effective
rank depends on sample count and can be inflated by noise, so we compare
it only within a shared evaluation set and interpret it together with seed
sensitivity. Finally, the six fixed corruption levels are single forward
passes, not a complete denoising trajectory. These probes motivate the
multi-level target construction. The end-to-end value of that target is
evaluated separately by the LIBERO supervision ablation in
Table~\ref{tab:target_ablation}.

\end{document}